%% file: main.tex
\documentclass{article}
\usepackage{iclr2027_conference,times}
\input{math_commands.tex}

\usepackage{hyperref}
\usepackage{url}
\usepackage{booktabs}
\usepackage{graphicx}
\usepackage{booktabs, graphicx, multirow, makecell, rotating}
\usepackage{amsmath,amssymb,amsthm,mathtools}

\usepackage{xcolor}
\usepackage{booktabs,multirow,rotating,graphicx}
\usepackage{colortbl}
\usepackage{subfig}
\usepackage{circledsteps}
\usepackage{wrapfig}
\usepackage{needspace}
\usepackage{float}
\usepackage{adjustbox}
\usepackage{threeparttable}
\usepackage{tcolorbox}
\usepackage{fontawesome}
\usepackage{pifont}
\usepackage{placeins}

\tcbset{
  theorybox/.style={
    colback=SeaGreen!10!CornflowerBlue!10,
    colframe=RoyalPurple!55!Aquamarine!100!,
    boxrule=0.5pt,
    arc=4pt,
    left=2pt,
    right=2pt,
    top=2pt,
    bottom=2pt
  }
}

\definecolor{uclablue}{rgb}{0.15, 0.45, 0.68}
\hypersetup{
    breaklinks,
    allcolors=black,
    colorlinks=true,
}

\title{Lift Out, Kick In: Decoupling Event Effects for Time Series Forecasting}

\author{
\begingroup
\setlength{\tabcolsep}{0pt}
\small
\begin{tabular}{@{}
  >{\centering\arraybackslash}p{0.220\textwidth}
  @{\hspace{0.033\textwidth}}
  >{\centering\arraybackslash}p{0.220\textwidth}
  @{\hspace{0.033\textwidth}}
  >{\centering\arraybackslash}p{0.220\textwidth}
  @{\hspace{0.033\textwidth}}
  >{\centering\arraybackslash}p{0.220\textwidth}@{}}
\textbf{Yunying Zhu}$^{\dagger}$\newline
  {\normalfont\small DiDi Chuxing, China}
& \textbf{Zichun Cao}$^{\dagger}$\newline
  {\normalfont\small DiDi Chuxing, China}
& \textbf{Siwei Sun}\newline
  {\normalfont\small DiDi Chuxing, China}
& \textbf{Xiaocen Guo}\newline
  {\normalfont\small DiDi Chuxing, China}
\\[12pt]
\textbf{Shengsheng Lin}$^{\ddagger}$\newline
  {\normalfont\small \makebox[\linewidth][c]{South China University}\newline
  of Technology, China}
& \textbf{Jing Hu}$^{\ddagger}$\newline
  {\normalfont\small \makebox[\linewidth][c]{South China University}\newline
  of Technology, China}
& \textbf{Enyun Yu}\newline
  {\normalfont\small DiDi Chuxing, China}
& \textbf{Zhichao Zou}$^{*}$\newline
  {\normalfont\small DiDi Chuxing, China}
\end{tabular}
\endgroup
}

\iclrfinalcopy
\renewcommand{\headrulewidth}{0pt}
\begin{document}
\begingroup
\setlength{\tabcolsep}{0pt}
\maketitle
\endgroup
\fancyhead{}
\begingroup
\renewcommand{\thefootnote}{\fnsymbol{footnote}}
\footnotetext[2]{Equal contribution.}
\footnotetext[3]{Work done while at DiDi Chuxing.}
\footnotetext[1]{Corresponding author. Email: \texttt{zouzhichao@didiglobal.com}}
\endgroup

\begin{abstract}

Real-world time series are often shaped by exogenous events whose effects are superimposed on the system’s underlying temporal dynamics. Historical observations therefore entangle persistent dynamics with event-induced deviations, making it difficult to determine which patterns should be extrapolated and which should be attributed to specific event contexts. Existing event-aware forecasting methods can incorporate event information or model event-related variation, but typically treat historical event attribution and future event generation separately. We propose Loki (Lift Out, Kick In), a general event-aware forecasting framework built on a shared Event Response Function (ERF). The ERF models event effects according to event type and relative temporal position and is shared across the historical and forecast windows. Loki lifts out estimated event-induced deviations from historical observations before forecasting, then uses the same ERF to kick in responses to known future events. This formulation separates event-response modeling from temporal forecasting while coupling historical attribution and future generation through a shared mechanism. Across five forecasting benchmarks, Loki achieves the best standardized MAE on all five datasets, reducing error by approximately \textcolor{black}{2.2\%--11.7\%} over the strongest non-Loki baseline. Ablations and controlled analyses further support the roles of Lift Out, Kick In, shared response modeling, and compatibility across forecasting backbones.

\end{abstract}

\section{Introduction}

\begin{wrapfigure}[23]{R}{0.45\textwidth}
\centering
\includegraphics[width=\linewidth]{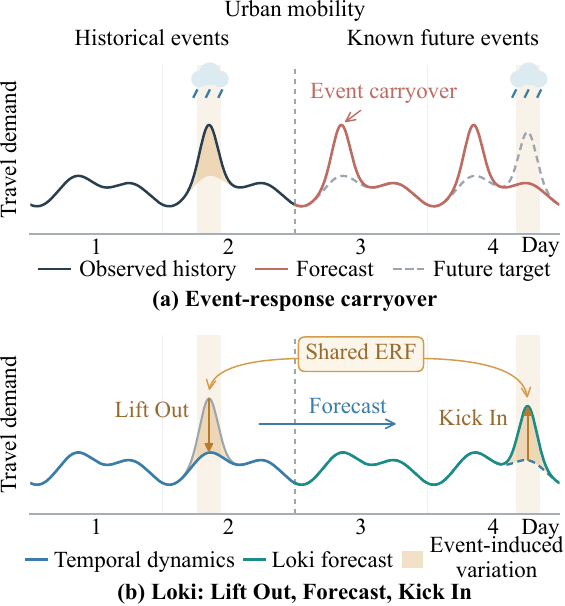}
\captionsetup{font=footnotesize,justification=raggedright,singlelinecheck=false}
\caption{Event-response carryover and Loki's Lift Out--Forecast--Kick In mechanism.}
\label{fig:intro-principle}
\end{wrapfigure}

\textcolor{black}{
In many real-world applications, time series exhibit recurring temporal patterns while also being influenced by external events such as holidays, promotions, and planned activities \citep{taylor2018prophet,rodrigues2019taxi,olivares2023nbeatsx}. These events can introduce deviations from regular patterns, complicating the extrapolation of historical observations into the future. A forecasting model may carry event-induced deviations into periods where they no longer apply, as illustrated in Figure~\ref{fig:intro-principle}(a), or fail to account for known upcoming events whose effects are absent from recent observations \citep{wolff2024spade}. This motivates modeling event-related variation separately from the underlying temporal dynamics.
}

\textcolor{black}{
A forecasting boundary separates observed from unobserved targets, but does not necessarily mark a change in how events affect them. When event effects are modeled separately from the underlying temporal dynamics, their estimates can be used to adjust historical observations before forecasting and to account for known events in future predictions. Moreover, recurring events of the same type may exhibit similar response patterns under comparable conditions \citep{taylor2018prophet,cooper1999promocast}. Although existing methods incorporate events through covariate conditioning, explicit effect modeling, or historical filtering \citep{lim2021tft,taylor2018prophet,wolff2024spade}, learning historical adjustment and future response generation independently does not explicitly exploit this potential commonality. This motivates investigating whether a shared response model can couple the two operations and improve event-response estimation.
}

\textcolor{black}{
We propose \textbf{Loki (Lift Out, Kick In)}, an event-aware forecasting framework built around a shared \textbf{Event Response Function (ERF)}. We focus on intermittently occurring, time-bounded events whose schedules are known across the historical and forecast windows. Their effects may extend beyond the event intervals themselves. As illustrated in Figure~\ref{fig:intro-principle}(b), Loki first \textbf{lifts out} estimated event-related variation from historical observations, allowing a forecasting backbone to predict the adjusted trajectory, and then \textbf{kicks in} the response associated with known future events. Using the same ERF for both operations allows their learning signals to jointly constrain a common response model through the forecasting objective. This formulation separates event-response modeling from temporal forecasting and can be combined with different forecasting backbones.
}

\textcolor{black}{
We evaluate Loki across five forecasting benchmarks spanning mobility, traffic, retail, store sales, and bike demand. Loki achieves the lowest standardized MAE among the compared methods on all five datasets. Structural ablations examine the contributions of explicit decomposition and shared response modeling, while controlled semi-synthetic experiments further evaluate the recovery of known event effects. Additional analyses demonstrate compatibility across forecasting backbones.
}

\textcolor{black}{
Our contributions are summarized as follows:
}
\begin{itemize}
\item \textcolor{black}{
\textbf{A shared event-response forecasting framework.}
We introduce Loki, which couples historical event adjustment and future response generation through a common learned ERF, using the Lift Out--Forecast--Kick In formulation.
}

\item \textcolor{black}{
\textbf{Empirical evaluation of shared response modeling.}
We demonstrate predictive competitiveness across diverse benchmarks and investigate the roles of explicit decomposition and response sharing through structural ablations, controlled event-effect recovery, and backbone compatibility analyses.
}
\end{itemize}

\section{Related Work}
\textcolor{black}{Time series forecasting methods differ in their use of available information, ranging from models relying solely on historical targets to those incorporating historical and known-future covariates. As Loki focuses on forecasting with event information, we review two related lines of research: methods that incorporate exogenous information through conditioning or fusion, and methods that explicitly model event-related variation as separate components.}

\subsection{Exogenous and Event-Conditioned Time Series Forecasting}

\textcolor{black}{Exogenous variables, including weather, holidays, and planned
activities, provide information beyond historical target observations.
Existing methods incorporate such information through different mechanisms.
TimeXer, CrossLinear, and XLinear model dependencies between endogenous
series and historical exogenous variables
\citep{wang2024timexer,zhou2025crosslinear,chen2026xlinear},
while TiDE, DAG, and TFT support known-future covariates through
conditioning, feature interaction, and variable selection
\citep{das2023tide,qiu2026dag,lim2021tft}.
These approaches incorporate external information without necessarily
representing event-induced variation as a separate response component.}

% 这样省空间，不要那个\section 对我删了 我感觉甚至不用分节了删过内容之后
% 这么长的内容还是要的 但现在2.1很短了 主要是2.2长 是不是可以每段分一个小标题黑体在前面
% 先压缩吧，到时候再讨论这个 ok
%\textbf{Exogenous and Event-Conditioned Time Series Forecasting.}
% Real-world time series are often influenced by external factors such as weather, holidays, promotions, and other contextual information, motivating extensive research on incorporating exogenous variables into forecasting models. Existing methods incorporate historical and known-future covariates through covariate conditioning, feature fusion, or explicit interaction mechanisms between endogenous and exogenous variables \citep{salinas2020deepar,lim2021tft,das2023tide,olivares2023nbeatsx,wang2024timexer}. \textcolor{black}{These approaches enable forecasting models to learn from external information, although event-related variation is not necessarily represented as a separate response component.}
% However, conditioning on external information does not by itself explicitly separate event-induced deviations in history from temporal dynamics to be extrapolated. This motivates a more explicit separation between event-associated variation and the temporal dynamics to be extrapolated.

%However, conditioning on event information does not by itself explicitly attribute which variations already present in the observed target trajectory were induced by past events. This motivates a more explicit separation between event-associated variation and the temporal dynamics to be extrapolated.

\subsection{Decomposition and Event-Effect Modeling}

\textcolor{black}{
Explicit modeling of event effects has a long history in time series forecasting. Intervention analysis characterizes dynamic responses to external interventions \citep{box1975intervention}, while dynamic regression and log-linear models represent exogenous and promotional effects through regression terms \citep{FOEKENS1994245}. Prophet models recurring holiday effects using shared event parameters \citep{taylor2018prophet}. In retail forecasting, \citet{wolters2021promotion} estimate seasonal demand from non-promotional periods and incorporate promotion and holiday effects through a multiplicative demand model. These approaches establish the foundations for separating event-related variation from regular temporal dynamics and reusing response patterns across event occurrences.
}

\textcolor{black}{
More recent methods explore different ways of structuring exogenous contributions and event-related variation within forecasting models. NBEATSx uses basis-expansion blocks to generate backcast and forecast components, with its interpretable configuration separating trend, seasonal, and exogenous contributions \citep{olivares2023nbeatsx}. SPADE filters historical event peaks and separately predicts future peaks using attention \citep{wolff2024spade}, while CEDAR refines an action-conditioned base forecast through event-driven residual correction using a two-stage procedure \citep{meng2026cedar}. More closely related to Loki, NeuralProphet jointly learns event components and autoregression, using shared coefficients at specified offsets relative to events \citep{triebe2021neuralprophet}. Its multiplicative event effects are scaled by the trend component rather than the full forecast. Together, these approaches demonstrate different ways to represent event-related variation and connect it to historical and future forecasting, differing in the response structures they learn and how these are applied.
}

\textcolor{black}{
Loki instead uses a shared, temporally structured Event Response Function (ERF) to model event responses across the history--forecast boundary. It uses temporal bases at multiple scales to model how the response unfolds over time, accounting for event intensity, position within the event, and differences across forecasting units. The same ERF is applied inversely to historical observations and multiplicatively to the full backbone forecast, coupling historical adjustment and future response generation through joint learning under the forecasting objective. Further details comparing Loki and related methods are provided in Appendix~\ref{app:method-comparison}.
}

\section{Method}
\label{sec:method}
% \subsection{Overview}
We consider event-aware forecasting with an observed target history
$X_{1:L}\in\mathbb{R}^{L\times F}$ and a future target
$Y_{L+1:L+H}\in\mathbb{R}^{H\times F}$, where $L$ and $H$ denote
the history and forecast lengths, respectively, and $F$ is the number of
target variables. The model is additionally given a known event schedule
$E_{1:T}\in\mathbb{R}^{T\times K}$ over both windows, where $T=L+H$,
and, when available, non-event contextual covariates
$C_{1:T}\in\mathbb{R}^{T\times D}$.

Loki formulates event-aware forecasting as a decomposition of the target
into an underlying trajectory and a structured event response:
\begin{equation}
\label{eq:target-decomposition}
Y_t = Y_t^0 \odot \exp(S_t),
\end{equation}
where $Y_t^0$ denotes the trajectory to be modeled by the forecasting
backbone and $S_t$ denotes the event response in log space. This formulation
provides a direct interpretation of the learned response in the original
target space: positive values of $S_t$ amplify the underlying trajectory,
negative values suppress it, and $S_t=0$ leaves the trajectory unchanged.
Moreover, responses from multiple event sources can be combined additively in
log space while corresponding to multiplicative effects on the original
target scale.

To model event effects consistently across the forecasting boundary, Loki
applies a shared Event Response Function (ERF) to the complete event schedule,
\begin{equation}
\label{eq:shared-erf}
S_{1:T} = \operatorname{ERF}_{\omega}(E_{1:T}),
\end{equation}
and partitions the resulting response into historical and future components,
$S^{\mathrm{hist}}$ and $S^{\mathrm{fut}}$.

% As illustrated in Figure~\ref{fig:loki-framework}, Loki follows three stages:
% \textbf{Lift Out}, which removes the estimated historical event response from
% the observed history; \textbf{Forecast}, which predicts the remaining temporal
% dynamics with a forecasting backbone; and \textbf{Kick In}, which restores the
% response associated with known future events. The same ERF therefore governs
% both historical event attribution and future event generation, coupling the
% two operations within a single forecasting framework.

\begin{figure}[!ht]
\centering
\includegraphics[width=\textwidth]{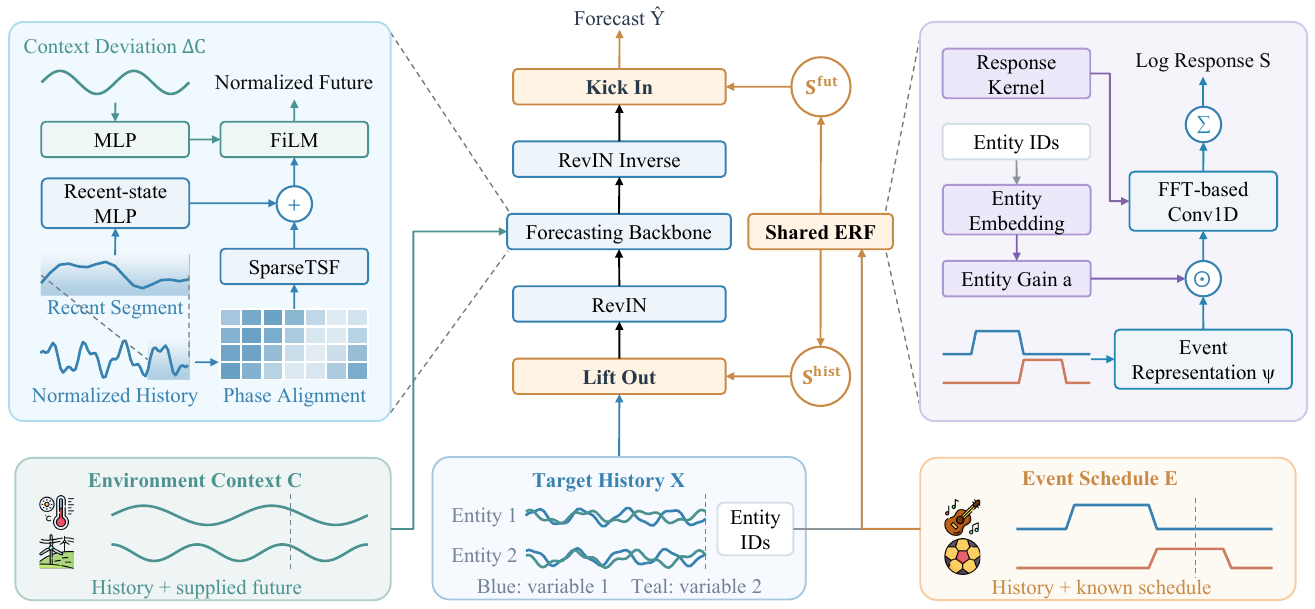}
\caption{\textbf{Overall architecture of Loki.} The center shows the
Lift Out--Forecast--Kick In pipeline, while the left and right panels detail
the forecasting backbone and the shared Event Response Function (ERF),
respectively.}
\label{fig:loki-framework}
\end{figure}

% \textcolor{black}{
As illustrated in Figure~\ref{fig:loki-framework}, Loki follows three stages:
\textbf{Lift Out}, which removes the estimated historical event response
$S^{\mathrm{hist}}$ from the observed history $X$ to obtain the event-adjusted
history $X^0$; \textbf{Forecast}, which predicts the underlying trajectory
$\widehat{Y}^0$ from $X^0$ using a forecasting backbone; and \textbf{Kick In},
which restores the future event response $S^{\mathrm{fut}}$ to obtain the final
forecast $\widehat{Y}$. The same ERF generates both $S^{\mathrm{hist}}$ and
$S^{\mathrm{fut}}$, coupling historical event attribution and future event
generation within a single forecasting framework.
% }

\subsection{Shared Event Response Function}

The shared Event Response Function (ERF) models how a known event induces a
structured response over time. For event channel $k$, forecasting unit $i$,
target variable $f$, and time $t$, the response depends on the event's
relative temporal position $r_{i,t,k}$, activity or intensity $d_{i,t,k}$,
and forecasting-unit-specific characteristics.

We first represent the event state using a multi-scale relative-position
encoding:
\begin{equation}
\label{eq:event-representation}
\psi_{i,t,k,m}
=
d_{i,t,k}\,\rho_m(r_{i,t,k}),
\end{equation}
where $m$ indexes the representation components and $\rho_m(\cdot)$ encodes the relative position within the event interval, as defined in Eq.~(\ref{eq:app-position-encoding}). Loki then models the
temporal response associated with each component using
\begin{equation}
\label{eq:response-kernel}
g_{k,m,f}(\tau)
=
\sum_{j=1}^{J}
W_{k,m,j,f}\,\beta_j(\tau),
\end{equation}
where $\{\beta_j\}$ is a fixed set of multi-scale bidirectional temporal
bases and $W$ contains the learned combination weights. The basis set
includes a zero-lag impulse, positive-lag exponential tails for persistent
post-event effects, and negative-lag exponential tails for anticipatory
effects. This bidirectional construction allows event influence to extend
beyond the event interval itself: responses may emerge before an event and
persist after it ends. Evaluating the same response structure across the
complete history--forecast window therefore captures both anticipatory and
carryover effects without introducing separate response models on the two
sides of the forecasting boundary. This parameterization allows events of the
same type to share a common response structure while learning target-specific
temporal profiles.

The response induced by event channel $k$ is
\begin{equation}
\label{eq:event-response}
s_{i,t,k,f}
=
\sum_m
a_{i,k,m}
\left(
g_{k,m,f} * \psi_{i,k,m}
\right)_t,
\end{equation}
where $a_{i,k,m}$ provides forecasting-unit-specific modulation, as defined in Eq.~(\ref{eq:app-entity-modulation}). Responses
from multiple event channels are aggregated additively in log space:
\begin{equation}
\label{eq:response-aggregation}
S_{i,t,f}
=
\sum_{k=1}^{K}
s_{i,t,k,f}.
\end{equation}

The shared ERF in Eq.~(\ref{eq:shared-erf}) is evaluated over the complete
observation-and-forecast window and then partitioned into
$S^{\mathrm{hist}}$ and $S^{\mathrm{fut}}$. Thus,
historical event attribution and future event generation share the same
response parameterization. Further details of the event representation,
temporal response bases, forecasting-unit modulation, and cross-boundary
computation are provided in
Appendices~\ref{app:event-representation}--\ref{app:cross-boundary}.

\subsection{Lift Out, Forecast, and Kick In}
Given the shared ERF, Loki applies Lift Out, forecasting, and Kick In
through Eqs.~(\ref{eq:lift-out})--(\ref{eq:kick-in}), respectively.
First, \textbf{Lift Out} removes the estimated historical event response from
the observed history:
\begin{equation}
\label{eq:lift-out}
\textcolor{black}{X^0 = X \odot \exp\!\left(-S^{\mathrm{hist}}\right).}
\end{equation}
This operation is intended to attribute event-associated variation to the ERF
before temporal extrapolation rather than to perform generic smoothing or
denoising. Window-level normalization (RevIN) \citep{kim2022revin} is then applied to the event-adjusted
history before forecasting.

Next, the forecasting backbone models the remaining temporal dynamics:
\begin{equation}
\label{eq:backbone-forecast}
\widehat{Y}^{0}
=
B_{\theta}(X^0,C),
\end{equation}
where $C$ denotes non-event contextual covariates when available. Loki
distinguishes bounded event variables, such as holidays, scheduled
activities, and weather episodes, from environmental covariates such as
temperature, precipitation, price, \textcolor{black}{and operating status}. The
former are modeled by the ERF, whereas the latter are provided to the
forecasting backbone as contextual inputs.

% In our main implementation, $B_{\theta}$ is a lightweight backbone based on
% SparseTSF \citep{lin2026sparsetsf} with an additional recent-state branch.
% Environmental context is incorporated through a lightweight FiLM-style
% modulation module \citep{perez2018film}. Detailed backbone and environmental
% modulation implementations are provided in
% Appendix~\ref{app:forecasting-learning}. These components can be replaced by
% other forecasting architectures without changing the event-response
% formulation.

% \textcolor{black}{
In our main implementation, $B_{\theta}$ is a lightweight backbone based on
SparseTSF \citep{lin2024sparsetsf,lin2026sparsetsf} with an additional recent-state branch.
The SparseTSF branch uses an MLP to model cross-period phase
relationships, while the recent-state branch uses an MLP to capture the most
recent seasonal segment. The outputs of the two branches are added to form the
baseline forecast. Environmental context is further incorporated through a
lightweight FiLM-style modulation module \citep{perez2018film}. These components
can be replaced by other forecasting architectures without changing the
event-response formulation.
% }

Finally, \textbf{Kick In} restores the response associated with known future
events:
\begin{equation}
\label{eq:kick-in}
\textcolor{black}{\widehat{Y} = \widehat{Y}^{0} \odot \exp\!\left(S^{\mathrm{fut}}\right).}
\end{equation}

Because $S^{\mathrm{hist}}$ and $S^{\mathrm{fut}}$ are generated by the same
ERF in Eq.~(\ref{eq:shared-erf}), historical event removal and future event generation are coupled
through a shared response model rather than learned as independent
corrections.

\subsection{Learning}

Loki is trained end-to-end using the final forecasting objective, so the
forecasting backbone and both uses of the shared ERF are optimized jointly.
Specifically, $S^{\mathrm{hist}}$ changes the event-adjusted trajectory
presented to the forecasting backbone, whereas $S^{\mathrm{fut}}$ directly
modifies the final prediction. The same forecasting objective therefore
constrains both historical attribution and future event generation through the
shared ERF. We use a scale-normalized mean absolute error:
\begin{equation}
\label{eq:forecast-loss}
\mathcal{L}
=
\frac{1}{BHF}
\sum_{b=1}^{B}
\sum_{h=1}^{H}
\sum_{f=1}^{F}
\left|
\frac{
\widehat{Y}_{b,h,f}
-
Y_{b,h,f}
}{
\sigma_f^{\mathrm{train}}
}
\right|,
\end{equation}
where $B$ denotes the batch size and $\sigma_f^{\mathrm{train}}$ denotes the
training-set scale of target $f$.

\textcolor{black}{The forecasting loss reaches the ERF directly through the future response in Kick In and indirectly through the backbone's dependence on the history adjusted by Lift Out. These two gradient contributions sum at the shared ERF parameters $\omega$ (Table~\ref{tab:forecast-gradient-paths} in Appendix~\ref{app:forecasting-learning}).}

We additionally use identity-preserving initialization: the event-response
branch is initialized to produce zero response, so Lift Out and Kick In
initially reduce to identity transformations. The environmental modulation
branch is initialized similarly, allowing optimization to start close to the
underlying forecasting backbone and gradually learn the additional
event-response structure. Further training and initialization details are
provided in Appendix~\ref{app:forecasting-learning}.

\section{Experiments}
\label{sec:experiments}
\textbf{Datasets.}
We evaluate Loki on five benchmarks spanning mobility, traffic, retail, store sales, and bike-rental demand: Ride-Hailing \citep{lin2026ridebench}, TaxiBJ \citep{zhang2017deepst}, M5 \citep{makridakis2022m5}, Rossmann \citep{kaggle2015rossmann}, and Seoul Bike \citep{uci2020seoulbike}. These datasets vary in temporal resolution, number of entities and targets, available covariates, and forecasting horizons. Detailed data splits, window construction, covariate definitions, and preprocessing are provided in Appendix~\ref{app:data-protocol}.

% \textbf{Baselines.}
% We compare Loki with eleven forecasting baselines: DLinear \citep{zeng2023dlinear}, PatchTST \citep{nie2023patchtst}, SparseTSF \citep{lin2026sparsetsf}, TimeMixer \citep{wang2024timemixer}, iTransformer \citep{liu2024itransformer}, TimeXer \citep{wang2024timexer}, CrossLinear \citep{zhou2025crosslinear}, XLinear \citep{chen2026xlinear}, TiDE \citep{das2023tide}, DAG \citep{qiu2026dag}, and TFT \citep{lim2021tft}. Models are evaluated under their supported information settings, including pure endogenous, historical-exogenous, and future-exogenous settings. Loki, TiDE, DAG, and TFT use future covariates without accessing future target values. We report standardized MAE and MSE, with scaling statistics computed solely from the training set and fixed across validation and test sets. Detailed protocols and configurations are provided in Appendices~\ref{app:forecasting-protocol} and~\ref{app:baseline-config}.

\textbf{Baselines.}
We compare Loki with \textcolor{black}{eighteen} forecasting baselines: DLinear \citep{zeng2023dlinear}, PatchTST \citep{nie2023patchtst}, SparseTSF \citep{lin2026sparsetsf}, TimeMixer \citep{wang2024timemixer}, iTransformer \citep{liu2024itransformer}, TimeXer \citep{wang2024timexer}, CrossLinear \citep{zhou2025crosslinear}, XLinear \citep{chen2026xlinear}, TiDE \citep{das2023tide}, DAG \citep{qiu2026dag}, and TFT \citep{lim2021tft}\textcolor{black}{, AutoARIMA-X \citep{hyndman2008autoarima}, a SCAN*PRO-style log-linear regression \citep{FOEKENS1994245}, Prophet \citep{taylor2018prophet}, NeuralProphet \citep{triebe2021neuralprophet}, NBEATSx \citep{olivares2023nbeatsx}, SPADE \citep{wolff2024spade}, and CEDAR \citep{meng2026cedar}}.
Models are evaluated under their supported information settings, including pure endogenous, historical-exogenous, and future-exogenous settings.
We report standardized MAE and MSE, with scaling statistics computed solely from the training set and fixed across validation and test sets.
Appendix~\ref{app:data-protocol} describes the evaluation protocol, and Appendix~\ref{app:experimental-implementation} details model training, checkpoint selection, and \textcolor{black}{baseline fitting procedures}.

\subsection{Main Results}
\input{table/main}

Table~\ref{tab:overall-performance} compares Loki against \textcolor{black}{eighteen} strong forecasting baselines across five heterogeneous benchmarks spanning mobility, traffic, retail, and bike-demand forecasting. Loki achieves the best standardized MAE on all five datasets, consistently outperforming the strongest non-Loki baseline by approximately \textcolor{black}{2.2\%--11.7\%}. \textcolor{black}{Loki also achieves the lowest standardized MSE on Ride-Hailing, TaxiBJ, M5, and Rossmann, and the second-lowest on Seoul Bike}, while attaining the best average rank across all compared methods.

\textcolor{black}{Loki also achieves the lowest MAE among models with access to known future covariates, including methods that explicitly model event effects.} Together with its consistent performance across datasets and evaluation metrics, this suggests that Loki's effectiveness extends across diverse forecasting domains rather than being driven by a particular benchmark or information setting. On the original data scale, Loki also achieves the lowest WMAPE on all five datasets (Table~\ref{tab:appendix-wmape}), further supporting its predictive advantage. Appendix~\ref{app:additional-metrics} reports the complete RMSE and WMAPE results.

% \subsection{Main Results}
% \input{table/main}
% Table~\ref{tab:overall-performance} compares Loki against eleven strong forecasting baselines across five heterogeneous benchmarks spanning mobility, traffic, retail, and bike-demand forecasting. Loki achieves the best standardized MAE on all five datasets, consistently outperforming the strongest non-Loki baseline by approximately 2.8\%--6.0\%. The gains extend beyond MAE: under standardized MSE, Loki ranks first on Ride-Hailing, TaxiBJ, and M5, and second on Rossmann and Seoul Bike, while attaining the best average rank across all compared methods. Importantly, Loki remains competitive with TiDE, DAG, and TFT under the same future-covariate setting, indicating that its gains are not attributable to an easier information regime.

% The consistent improvements across datasets and evaluation metrics suggest that Loki's effectiveness generalizes across diverse forecasting domains rather than being driven by a particular benchmark or metric. On the original data scale, Loki also achieves the lowest WMAPE on all five datasets (Table~\ref{tab:appendix-wmape}), further supporting its predictive advantage. Appendix~\ref{app:additional-metrics} reports the complete RMSE and WMAPE results.

\subsection{Ablation Study}
\input{table/ablation}

We first examine the value of explicit Lift Out/Kick In decomposition. \textbf{Direct Fusion} uses the same event information but incorporates it directly into the forecaster, without separating event responses from regular temporal dynamics. \textcolor{black}{Full Loki achieves lower MAE on all five datasets (Table~\ref{tab:ablation}). To isolate the role of the response function within this decomposition, \textbf{Event Lookup} retains Lift Out and Kick In but replaces the ERF with a shared, learnable lag-based response table. Full Loki achieves lower MAE on four of five datasets.}

The two one-sided variants, \textbf{History-only} and \textbf{Future-only}, retain only Lift Out and Kick In, respectively. \textcolor{black}{Full Loki has lower MAE than History-only on four of five datasets and lower MAE than Future-only on all five.}

\textbf{Independent ERF} retains the same two-sided structure but learns separate response functions for the historical and forecast windows. \textcolor{black}{The shared ERF achieves lower MAE than Independent ERF on four of five datasets. Appendix~\ref{app:capacity-ablation} provides additional comparisons with matched parameter budgets.}

\textcolor{black}{Finally, \textbf{Additive Loki} uses the same backbone, shared ERF architecture, and training protocol. It subtracts the estimated event effect from historical observations and adds the future effect to the backbone forecast. The multiplicative formulation achieves lower MAE on all five datasets. Appendix~\ref{app:additive-ablation} reports the means and standard deviations across seeds.}

% \subsection{Ablation Study}
% \input{table/ablation}
% As shown in Table~\ref{tab:ablation}, Loki consistently outperforms the parameter-matched \textbf{Direct Fusion} variant, which uses the same event information but injects it directly into the forecasting pathway without explicit Lift Out/Kick In decomposition. The improvement holds across both datasets and metrics and is particularly substantial on M5, with reductions of 6.52\% in MAE and 15.13\% in MSE. This controlled comparison suggests that the gains arise from the proposed decomposition rather than event conditioning alone.

% Loki also consistently outperforms both one-sided variants, \textbf{History-only} and \textbf{Future-only}, which retain only Lift Out and Kick In, respectively. On M5, Loki improves over the stronger one-sided variant by 4.76\% in MAE and 13.42\% in MSE, demonstrating that jointly modeling historical and future event effects is important for capturing the full response pattern.

% Loki outperforms \textbf{Independent ERF}, which replaces the shared ERF with two architecturally identical but independently parameterized response functions for the historical and forecast windows. Despite introducing additional response parameters, Independent ERF performs consistently worse across both datasets and metrics. On M5, the shared ERF yields 5.41\% lower MAE and 6.86\% lower MSE, supporting the effectiveness of parameter sharing for learning a unified response function across historical and future windows.

\subsection{Further Analyses}

\subsubsection{Controlled Event-Effect Recovery}

Forecasting accuracy alone does not establish whether the learned ERF
correctly recovers event-associated variation. To examine this directly,
we construct semi-synthetic versions of Ride-Hailing and M5 by injecting
known event responses into the original reference series while
preserving their underlying temporal dynamics and noise.
\textcolor{black}{We consider both multiplicative and additive injection
mechanisms.}
Models observe the corresponding artificial event descriptors but
receive no direct supervision on the injected responses.
We measure historical restoration against the pre-injection series and
future event-effect recovery against the known injected increments,
using standardized MAE in both cases.
\textcolor{black}{Future effects from different response formulations
are compared in the same target-increment space.}
Full experimental details are provided in
Appendix~\ref{app:controlled-recovery}.

\input{table/controlled_recovery}

\textbf{Recovery and parameter sharing.}
As shown in Table~\ref{tab:controlled-recovery}, under multiplicative
injection, Full Loki reduces both historical restoration and future
event-effect MAE by over 93\% relative to no-recovery reference
across Ride-Hailing and M5.
\textcolor{black}{Compared with Independent ERFs, the shared
formulation further reduces historical restoration MAE by 29.6\%
and 56.6\% on the two datasets, respectively, while also improving
future event-effect recovery. These results support the benefit of
coupling historical attribution and future response generation through
a common response model.}

\textcolor{black}{\textbf{Sensitivity to response form.}
Table~\ref{tab:controlled-recovery} also shows that recovery accuracy
depends on the response formulation. Under multiplicative injection,
Full Loki consistently outperforms Additive Loki across both datasets
and recovery metrics. Conversely, under additive injection, the
additive formulation achieves substantially lower errors. For example,
future event-effect MAE on Ride-Hailing decreases from 0.1004 with
multiplicative Loki to 0.0247 with Additive Loki. While the
multiplicative formulation still recovers part of the additive effects,
the results demonstrate the importance of matching the response
formulation to the underlying event mechanism.}

\textcolor{black}{Additional semi-synthetic experiments in
Appendix~\ref{app:state-dependent-recovery} show that Loki also improves
historical restoration and future effect recovery when response intensity
depends on the recent reference-series level, although recovery remains
imperfect.}

\textbf{Null control.}
Finally, we examine spurious attribution by retaining the artificial
event descriptors without injecting corresponding effects.
Full Loki produces low historical restoration and future event-effect
errors across both datasets,
\textcolor{black}{with all four MAE values below 0.01.}
This complements the recovery results by showing that the learned
ERF introduces limited spurious responses when no artificial effect
is present.

\Needspace{23\baselineskip}
\subsubsection{Event-Regime Analysis}
\input{table/across}

We partition test windows according to whether active discrete-event indicators appear in the historical and forecast windows. Quiet (Q) denotes no active indicator, while Event (E) denotes at least one, yielding four history-to-forecast regimes: Q$\rightarrow$Q, E$\rightarrow$Q, Q$\rightarrow$E, and E$\rightarrow$E. As shown in Figure~\ref{fig:event-regime-analysis}(a), Loki's relative MAE reduction becomes larger as event involvement increases, rising from \textcolor{black}{1.75\%} in Q$\rightarrow$Q to \textcolor{black}{2.38\%} in E$\rightarrow$Q, \textcolor{black}{3.39\%} in Q$\rightarrow$E, and \textcolor{black}{6.25\%} in E$\rightarrow$E. This pattern suggests that Loki's gains are not uniform across samples, but become more pronounced in regimes where event-aware modeling is more relevant.

Figure~\ref{fig:event-regime-analysis}(b) further separates the contributions of the two components. Kick In, measured by Loki's additional gain over History-only, contributes more when the forecast window is event-active, reaching \textcolor{black}{6.05\%} in Q$\rightarrow$E and \textcolor{black}{7.92\%} in E$\rightarrow$E, compared with \textcolor{black}{4.76\%} in Q$\rightarrow$Q and \textcolor{black}{5.63\%} in E$\rightarrow$Q. This is consistent with its role in modeling responses associated with future events. Lift Out, measured relative to Future-only, has its largest incremental contribution in E$\rightarrow$Q (\textcolor{black}{0.82\%}), where event-affected history transitions into an event-quiet future and historical event-related variation is less directly transferable to the forecast horizon. Overall, the regime-dependent patterns align with the intended division of roles between Kick In and Lift Out across the two sides of the forecasting boundary.

% Keep Table 4 from floating above the preceding analysis on this page.
\suppressfloats[t]
\input{table/backbone}
% \textbf{Backbone Compatibility.}
\subsubsection{Backbone Compatibility}

Loki decouples event-response modeling from temporal forecasting, allowing its Event Response Function to plug into different forecasting backbones. To evaluate whether the proposed event-response mechanism depends on the default forecasting architecture, we instantiate Loki with three representative backbones: SparseTSF, PatchTST, and TimeMixer. These models differ substantially in their temporal modeling strategies, providing a controlled test of whether Loki’s gains persist beyond a particular backbone design. We evaluate all three variants on Ride-Hailing and M5.

As shown in Table~\ref{tab:backbone_compatibility}, Loki reduces MSE across all six backbone--dataset combinations and MAE in five, with essentially unchanged MAE for PatchTST on Ride-Hailing. MSE reductions range from 4.40\% to 10.26\% on Ride-Hailing and from 7.26\% to 11.76\% on M5. These results support the compatibility of Loki's event-response modeling with diverse forecasting backbones on the two evaluated datasets.

\FloatBarrier
\section{Limitations}
\textcolor{black}{Loki is designed for forecasting settings where event schedules are available and event responses exhibit reusable patterns across occurrences. Its shared ERF models these patterns through event and entity information, while the multiplicative Lift Out/Kick In formulation represents event effects as relative changes to the underlying trajectory. These modeling choices couple historical adjustment and future response generation through a common response rule, while limiting the ERF's freedom to absorb ordinary temporal variation that should be modeled by the forecasting backbone. This constraint also defines the model's scope: incomplete or inaccurate event schedules, responses that depend strongly on unmodeled conditions, or effects that depart substantially from a proportional form may be less accurately represented.}

% Loki assumes that relevant event descriptors are available for both historical and forecast windows. While suitable for scheduled or externally observed events, this assumption does not cover latent, unrecorded, or previously unseen events whose occurrence or semantics are unavailable to the model. The shared Event Response Function (ERF) further assumes that response patterns learned from historical events remain informative for future ones, which may be less appropriate under substantial shifts in event effects, unseen event types, or changing entity-specific responses. Extending the response model to adapt to distribution shifts or previously unseen events is an important direction for more non-stationary settings.

The learned ERF should be interpreted as a predictive decomposition of event-associated variation rather than a causal effect estimator. In observational data, the underlying dynamics and event-induced component are generally not separately identifiable without additional assumptions or interventions. Controlled or semi-synthetic experiments can evaluate recovery of known injected event components by construction, but such recovery does not establish causal identifiability in real-world settings.

\section{Conclusion}
%We introduced Loki, an event-aware forecasting framework that explicitly separates event effects from underlying temporal dynamics. Historical observations may already contain event-induced deviations, while the events present in the forecast horizon may differ from those in history. Without explicitly separating these effects, a forecasting model may carry historical event-induced variation into the future even when it should not persist, while failing to account for event responses that should emerge in the forecast horizon. Loki addresses this mismatch by using a shared Event Response Function to Lift Out event-induced variation from the historical observations before backbone forecasting, and then Kick In the corresponding event responses in the forecast horizon.

\textcolor{black}{The forecasting boundary separates observed history from future targets, but does not necessarily mark a change in how events affect them. Motivated by this observation, we introduced Loki, an event-aware forecasting framework built around a shared, temporally structured Event Response Function (ERF). The ERF models responses according to event characteristics and relative temporal position, allowing event-associated variation to be lifted out from historical observations and responses to known future events to be kicked in after forecasting. Both operations jointly constrain the shared ERF through the final forecasting objective, while event-response modeling remains distinct from temporal forecasting. Across five forecasting benchmarks, Loki achieves the lowest standardized MAE among the compared methods. Structural ablations support the contributions of explicit decomposition, response parameterization, and sharing, while controlled experiments further demonstrate the recovery of known event effects and the benefits of shared response learning. Additional experiments show compatibility across different forecasting backbones. Together, these findings support jointly learning a shared event-response model as a useful inductive bias for event-aware time series forecasting.}

% In this paper, we introduced Loki, a general framework for event-aware time series forecasting that separates event-induced variation from the temporal dynamics to be extrapolated. Loki uses a shared Event Response Function across the forecasting boundary: event-related deviations are lifted out from historical observations before temporal forecasting, while the same response mechanism generates and kicks in the effects associated with known future events. This formulation couples historical attribution and future generation while keeping event-response modeling distinct from the forecasting backbone. Experiments show that this separation improves forecasting performance, and mechanism-oriented analyses consistently support the intended Lift Out/Kick In interpretation. The learned ERF recovers controlled event responses, Loki provides larger gains in event-relevant regimes, and the same mechanism remains effective across different forecasting backbones. Overall, our results suggest that event effects are better treated as a shared response process across history and future, rather than being absorbed into the underlying temporal dynamics or modeled separately on the two sides of the forecasting boundary.

\section*{AI Use Statement}

Generative AI tools were used in this work for the following tasks: literature exploration and summarization, refinement of the conceptual framing, feedback on research methodology and experimental design, interpretation of experimental results, manuscript structuring, and drafting and editing portions of the paper for clarity and readability.

Generative AI tools were not used for the following disclosure-relevant tasks: generating synthetic datasets, implementing the proposed method, formulating mathematical claims, or developing or verifying formal proofs.

The authors reviewed and verified all AI-assisted outputs by checking cited references against the original sources, independently reviewing methodological and experimental suggestions, validating all reported experimental results against the underlying experiment outputs, and manually reviewing and revising all manuscript text. The authors take full responsibility for all text, claims, code, results, and other artifacts contained in this work.

% \section*{Ethics Statement}
\section*{Reproducibility Statement}

The main paper describes Loki's architecture, learning objective, and evaluation setting. Appendix~\ref{app:method-details} provides implementation details for the shared ERF and forecasting backbone. Appendix~\ref{app:data-protocol} describes dataset construction, covariates, preprocessing, and the chronological splitting and windowing protocols. Appendix~\ref{app:experimental-implementation} documents the computational environment, model training, \textcolor{black}{checkpoint selection, and baseline fitting procedures}. Appendix~\ref{app:additional-results} defines the additional evaluation metrics and explains how variability across training seeds is reported. Appendix~\ref{app:controlled-recovery} describes how synthetic event effects are injected, how recovery is evaluated, and \textcolor{black}{how results are aggregated across training seeds and generated configurations}.
% \section*{Author Contributions}
% \section*{Acknowledgments}

\bibliography{references}
\bibliographystyle{iclr2027_conference}

\appendix
\numberwithin{equation}{section}
\input{Appendix}
\end{document}

%% file: math_commands.tex
\usepackage{amsmath,amsfonts,bm}

\def\eqref#1{equation~\ref{#1}}
\def\1{\bm{1}}

\DeclareMathAlphabet{\mathsfit}{\encodingdefault}{\sfdefault}{m}{sl}
\SetMathAlphabet{\mathsfit}{bold}{\encodingdefault}{\sfdefault}{bx}{n}

%% file: table/main.tex
\begin{table*}[!htb]
\caption{Forecasting performance across five datasets. Models are grouped by covariate availability and event-handling mechanism. The best results are highlighted in \textbf{bold} and the second best are \underline{underlined}.}
\label{tab:overall-performance}
\centering
\small
\setlength{\tabcolsep}{4pt}
\resizebox{\textwidth}{!}{%
\begin{tabular}{l|cc|cc|cc|cc|cc}
\toprule
Dataset & \multicolumn{2}{c|}{Ride-Hailing} & \multicolumn{2}{c|}{TaxiBJ} & \multicolumn{2}{c|}{M5} & \multicolumn{2}{c|}{Rossmann} & \multicolumn{2}{c}{Seoul Bike} \\
\cmidrule(lr){2-3}\cmidrule(lr){4-5}\cmidrule(lr){6-7}\cmidrule(lr){8-9}\cmidrule(lr){10-11}
Metrics & MAE & MSE & MAE & MSE & MAE & MSE & MAE & MSE & MAE & MSE \\
\midrule
\multicolumn{11}{l}{\textit{Endogenous-only}} \\
DLinear & 0.1890 & 0.2271 & 0.1264 & 0.0563 & 0.1068 & 0.0309 & 0.3352 & 0.2702 & 0.6722 & 0.8312 \\
PatchTST & \underline{0.1618} & 0.2171 & 0.1323 & 0.0698 & 0.1022 & 0.0298 & 0.2482 & 0.2069 & 0.5343 & 0.7825 \\
SparseTSF & 0.1804 & 0.2265 & 0.1357 & 0.0645 & 0.1026 & 0.0298 & 0.2753 & 0.2439 & 0.5616 & 0.6540 \\
TimeMixer & 0.1653 & 0.2212 & 0.1311 & 0.0677 & 0.1025 & 0.0304 & 0.2490 & 0.2019 & 0.5047 & 0.6679 \\
iTransformer & 0.1670 & 0.2208 & 0.1310 & 0.0669 & 0.1037 & 0.0314 & 0.2507 & 0.2028 & 0.5547 & 0.7451 \\
\midrule
\multicolumn{11}{l}{\textit{Historical covariates}} \\
TimeXer & 0.1713 & 0.2232 & 0.1324 & 0.0612 & 0.1068 & 0.0321 & 0.2529 & 0.2018 & 0.5204 & 0.6860 \\
CrossLinear & 0.1649 & 0.2222 & 0.1276 & 0.0619 & 0.1023 & 0.0294 & 0.2551 & 0.2037 & 0.5050 & 0.7205 \\
XLinear & 0.1659 & 0.2193 & 0.1391 & 0.0723 & \underline{0.1019} & 0.0304 & 0.2518 & 0.1973 & 0.5097 & 0.7058 \\
\midrule
\multicolumn{11}{l}{\textit{Known-future covariates: events as inputs}} \\
TiDE & 0.1625 & \underline{0.2008} & 0.1333 & 0.0597 & 0.1080 & 0.0309 & 0.1597 & 0.0600 & 0.6079 & 0.7618 \\
DAG & 0.1655 & 0.2198 & 0.1296 & 0.0629 & 0.1059 & 0.0317 & 0.1554 & \underline{0.0551} & 0.4747 & 0.5804 \\
TFT & 0.1692 & 0.2113 & \underline{0.1191} & 0.0538 & 0.1067 & 0.0334 & 0.1591 & 0.0669 & \underline{0.3310} & \textbf{0.2076} \\
\midrule
\multicolumn{11}{l}{\textit{Known-future covariates: explicit event-effect decomposition}} \\
AutoARIMA-X & 0.4607 & 29.3990 & 0.1718 & 0.1058 & 0.1179 & 0.0442 & 0.1927 & 0.0775 & 0.5843 & 0.5944 \\
Log-linear (SCAN*PRO) & 0.2098 & 0.2464 & 0.1214 & \underline{0.0503} & 0.1450 & 0.0728 & 0.1779 & 0.0802 & 0.3585 & 0.2860 \\
Prophet & 0.2602 & 0.6072 & 0.2040 & 0.1287 & 0.1083 & 0.0333 & 0.1920 & 0.0806 & 0.7736 & 1.0910 \\
NeuralProphet & 0.1882 & 0.2389 & 0.1323 & 0.0656 & 0.1032 & \underline{0.0287} & 0.2045 & 0.0941 & 1.2371 & 3.1938 \\
NBEATSx & 0.1727 & 0.2266 & 0.1282 & 0.0618 & 0.1059 & 0.0306 & 0.1768 & 0.0710 & 0.5159 & 0.6080 \\
\midrule
\multicolumn{11}{l}{\textit{Known-future covariates: event filtering / residual correction}} \\
SPADE & 0.1821 & 0.2296 & 0.1316 & 0.0627 & 0.1173 & 0.0387 & 0.1718 & 0.0749 & 0.4497 & 0.4235 \\
CEDAR & 0.2252 & 0.2974 & 0.1247 & 0.0636 & 0.1093 & 0.0355 & \underline{0.1545} & 0.0628 & 0.3578 & 0.2949 \\
\midrule
\multicolumn{11}{l}{\textit{Known-future covariates: shared event response}} \\
Loki & \textbf{0.1570} & \textbf{0.1979} & \textbf{0.1128} & \textbf{0.0475} & \textbf{0.0958} & \textbf{0.0250} & \textbf{0.1365} & \textbf{0.0474} & \textbf{0.3238} & \underline{0.2473} \\
\bottomrule
\end{tabular}%
}
\end{table*}

%% file: table/ablation.tex
\begin{table*}[t]
\caption{Ablation study of Loki across five datasets. Standardized MAE and MSE are reported (lower is better). The best results are highlighted in \textbf{bold}.}
\label{tab:ablation}
\centering
\small
\setlength{\tabcolsep}{3.5pt}
\resizebox{\textwidth}{!}{%
\begin{tabular}{l|cc|cc|cc|cc|cc}
\toprule
Dataset
& \multicolumn{2}{c|}{Ride-Hailing}
& \multicolumn{2}{c|}{TaxiBJ}
& \multicolumn{2}{c|}{M5}
& \multicolumn{2}{c|}{Rossmann}
& \multicolumn{2}{c}{Seoul Bike} \\
\cmidrule(lr){2-3}
\cmidrule(lr){4-5}
\cmidrule(lr){6-7}
\cmidrule(lr){8-9}
\cmidrule(lr){10-11}
Metrics
& MAE & MSE
& MAE & MSE
& MAE & MSE
& MAE & MSE
& MAE & MSE \\
\midrule

Only Backbone
& 0.1636 & 0.2175
& 0.1228 & 0.0592
& 0.1035 & 0.0307
& 0.1871 & 0.0830
& 0.3349 & 0.2628 \\

Direct Fusion
& 0.1612 & 0.2129
& 0.1227 & 0.0594
& 0.1026 & 0.0301
& 0.1817 & 0.0796
& 0.3246 & 0.2425 \\

Event Lookup
& 0.1644 & 0.2028
& 0.1591 & 0.1346
& 0.0988 & 0.0272
& 0.1469 & 0.0536
& 0.3207 & 0.2393 \\

History-only
& 0.1680 & 0.2198
& 0.1180 & 0.0531
& 0.1084 & 0.0356
& 0.1890 & 0.0843
& \textbf{0.3151} & \textbf{0.2351} \\

Future-only
& 0.1578 & 0.1992
& 0.1802 & 0.1425
& 0.0985 & 0.0280
& 0.1476 & 0.0540
& 0.3267 & 0.2458 \\

Independent ERF
& 0.1572 & \textbf{0.1975}
& 0.1189 & 0.0524
& 0.0983 & 0.0276
& 0.1396 & 0.0502
& 0.3223 & 0.2435 \\

\textcolor{black}{Additive Loki}
& \textcolor{black}{0.1579} & \textcolor{black}{0.2035}
& \textcolor{black}{0.1313} & \textcolor{black}{0.0668}
& \textcolor{black}{0.0966} & \textcolor{black}{0.0253}
& \textcolor{black}{0.1590} & \textcolor{black}{0.0608}
& \textcolor{black}{0.3362} & \textcolor{black}{0.2619} \\

\midrule
\textit{Full Loki}
& \textbf{0.1570} & 0.1979
& \textbf{0.1128} & \textbf{0.0475}
& \textbf{0.0958} & \textbf{0.0250}
& \textbf{0.1365} & \textbf{0.0474}
& 0.3238 & 0.2473 \\

\bottomrule
\end{tabular}%
}
\end{table*}

%% file: table/controlled_recovery.tex
\begin{table}[!b]
\caption{Controlled event-effect recovery under multiplicative,
additive, and null injections on Ride-Hailing and M5.
History restoration and future event-effect errors are
reported as standardized MAE (lower is better).
Multiplicative responses are converted to target increments
using the pre-injection reference series for evaluation.
The null setting measures spurious attribution when no
artificial effect is injected. Best results within each
non-null injection setting are highlighted in bold.}
\label{tab:controlled-recovery}
\centering
\scriptsize
\resizebox{\linewidth}{!}{%
\begin{tabular}{llcccc}
\toprule
 & & \multicolumn{2}{c}{Ride-Hailing}
   & \multicolumn{2}{c}{M5} \\
\cmidrule(lr){3-4}\cmidrule(lr){5-6}
Injection & Method
 & History & Future effect 
 & History & Future effect  \\
\midrule
\multirow{4}{*}{Multiplicative}
 & No recovery (reference)
 & 0.3333 & 0.3284 & 0.1927 & 0.2493 \\
 & Independent ERFs
 & 0.0294 & 0.0245 & 0.0272 & 0.0173 \\
 & Full Loki
 & \textbf{0.0207} & \textbf{0.0228}
 & \textbf{0.0118} & \textbf{0.0155} \\
 & Additive Loki
 & 0.1149 & 0.1135 & 0.0469 & 0.0675 \\
\midrule
\multirow{3}{*}{Additive}
 & No recovery (reference)
 & 0.3176 & 0.3138 & 0.1825 & 0.2003 \\
 & Full Loki
 & 0.0782 & 0.1004 & 0.0319 & 0.0552 \\
 & Additive Loki
 & \textbf{0.0267} & \textbf{0.0247}
 & \textbf{0.0134} & \textbf{0.0141} \\
\midrule
Null
 & Full Loki
 & 0.0097 & 0.0082 & 0.0052 & 0.0062 \\
\bottomrule
\end{tabular}%
}
\end{table}

%% file: table/across.tex
\begin{wrapfigure}{r}{0.60\textwidth}
  \vspace{-\intextsep} % Align image top with the first text line.
  \centering
  \includegraphics[width=\linewidth]{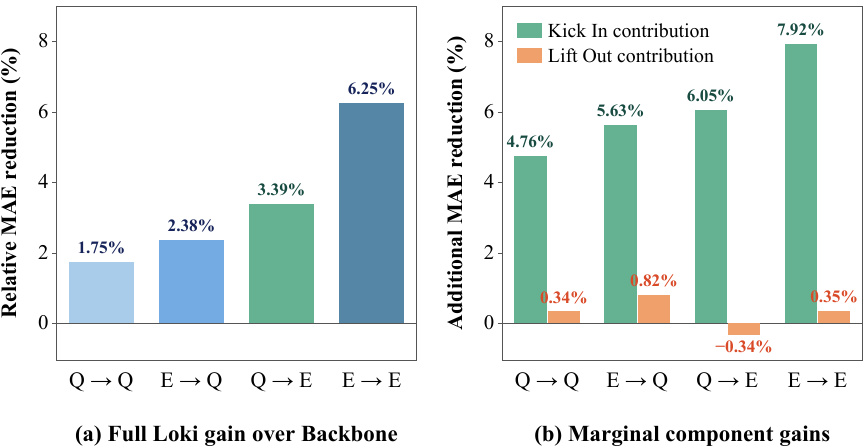}
  \caption{
  \textbf{Event-regime analysis on Ride-hailing.}
  Regimes are denoted as history$\rightarrow$forecast, with Q and E indicating quiet and event-active windows. (a) Relative MAE reduction of Full Loki over the Backbone.
  (b) Marginal Kick In and Lift Out contributions, measured against History-only and Future-only, respectively.
  }
  \label{fig:event-regime-analysis}
\end{wrapfigure}

%% file: table/backbone.tex
\begin{table}[t]
  \centering
  \caption{Compatibility of Loki with different forecasting backbones. Means over five training seeds; improvements use unrounded means.
  }
  \label{tab:backbone_compatibility}
  \small
  \begin{tabular*}{\linewidth}{@{\extracolsep{\fill}}ccccc@{}}
  \toprule
  Dataset & \multicolumn{2}{c}{Ride-Hailing} & \multicolumn{2}{c}{M5} \\
  \cmidrule(lr){2-3}\cmidrule(lr){4-5}
  Metrics & MAE & MSE & MAE & MSE \\
  \midrule
  PatchTST & 0.1618 & 0.2171 & 0.1022 & 0.0298 \\
  +Loki & 0.1618 & \textbf{0.2021} & \textbf{0.0988} & \textbf{0.0264} \\
  Improve & -0.005\% & 6.90\% & 3.35\% & 11.49\% \\
  \midrule
  SparseTSF & 0.1804 & 0.2265 & 0.1026 & 0.0298 \\
  +Loki & \textbf{0.1705} & \textbf{0.2033} & \textbf{0.0979} & \textbf{0.0263} \\
  Improve & 5.50\% & 10.26\% & 4.64\% & 11.76\% \\
  \midrule
  TimeMixer & 0.1653 & 0.2212 & 0.1025 & 0.0304 \\
  +Loki & \textbf{0.1651} & \textbf{0.2114} & \textbf{0.1015} & \textbf{0.0282} \\
  Improve & 0.12\% & 4.40\% & 1.05\% & 7.26\% \\
  \bottomrule
  \end{tabular*}
\end{table}

%% file: Appendix.tex
\newpage

\section{Additional Method Details}
\label{app:method-details}

\subsection{Event Representation}
\label{app:event-representation}

Loki represents each event through three complementary aspects: its \textbf{sustained activity}, its \textbf{proximity to the event onset}, and its \textbf{proximity to the event end}. Historical and future event covariates are concatenated and processed jointly over the complete $L+H$ window before the response is split into historical and future components.

Let $e_{i,t,k}$ denote the value of event channel $k$ for forecasting unit $i$ in the event schedule $E$ at time $t$. For continuous or flag-valued event channels, nonzero values are first normalized using the training-set nonzero RMS statistic $\mathrm{RMS}^{\mathrm{train}}_k$,
\begin{equation}
\label{eq:app-dose-normalization}
\bar e_{i,t,k}
=
\frac{e_{i,t,k}}{\mathrm{RMS}^{\mathrm{train}}_k}.
\end{equation}
The activity or intensity variable $d_{i,t,k}$ used in the main text is then obtained through a learnable event-type-specific power transform,
\begin{equation}
\label{eq:app-dose-transform}
d_{i,t,k}
=
\operatorname{sign}(\bar e_{i,t,k})
|\bar e_{i,t,k}|^{\gamma_k},
\qquad
\gamma_k=\exp(\eta_k).
\end{equation}

Consecutive nonzero values in the same event channel are treated as one event interval. Let $p_{i,t,k}$ and $q_{i,t,k}$ denote the distances from the beginning and end of that interval, respectively, and let $v^p_{i,t,k}$ and $v^q_{i,t,k}$ indicate whether the corresponding boundary is observable within the current window. These quantities instantiate the relative-position state $r_{i,t,k}$ introduced in the main text. We define its multi-scale encoding directly as
\begin{equation}
\label{eq:app-position-encoding}
\rho(r_{i,t,k})
=
\left[
1,\,
\left\{e^{-p_{i,t,k}/s}v^p_{i,t,k}\right\}_{s\in\mathcal S_{\mathrm{pos}}},\,
\left\{e^{-q_{i,t,k}/s}v^q_{i,t,k}\right\}_{s\in\mathcal S_{\mathrm{pos}}}
\right],
\end{equation}
where
\[
\mathcal S_{\mathrm{pos}}
=
\operatorname{unique}
\left\{
\text{steps-per-day},
\text{season-length}
\right\}.
\]

Here $\rho_m(r_{i,t,k})$ denotes the $m$-th component of the vector $\rho(r_{i,t,k})$. The final event representation is therefore
\[
\psi_{i,t,k,m}
=
d_{i,t,k}\rho_m(r_{i,t,k}).
\]

The position encoder itself describes the event interval; anticipatory and post-event effects are produced by the bidirectional temporal response function below.

\subsection{Shared Event Response Function}
\label{app:shared-erf}

\begin{center}
\begin{minipage}{\linewidth}
\centering
\captionsetup{type=figure}
\includegraphics[width=0.8\linewidth]{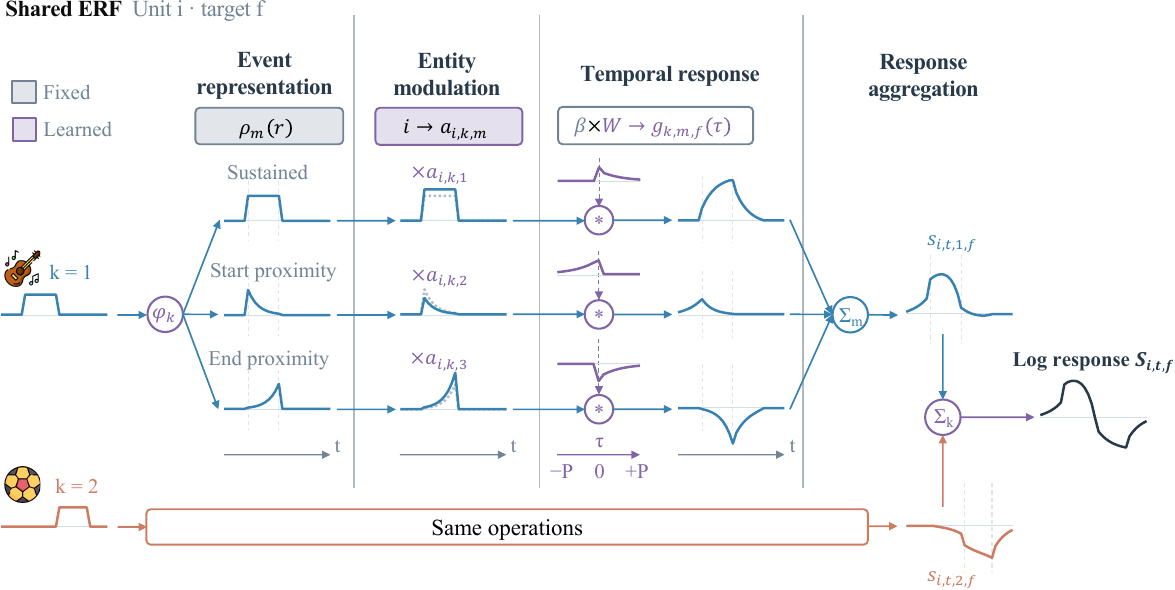}
\caption{Computation of the shared Event Response Function (ERF).}
\label{fig:appendix-erf}
\end{minipage}
\end{center}

Figure~\ref{fig:appendix-erf} summarizes the computation of the shared ERF from event representation through response aggregation. The ERF uses a compact fixed basis bank to represent \textbf{instantaneous}, \textbf{anticipatory}, and \textbf{post-event carryover} event responses. For response support $P$, the basis contains a zero-lag impulse,
\[
\beta_0(\tau)=\mathbf{1}[\tau=0],
\]
together with positive- and negative-lag exponential tails,
\begin{equation}
\label{eq:app-temporal-bases}
\beta_s^+(\tau)
=
\frac{
\mathbf{1}[\tau>0]e^{-\tau/s}
}{
\sum_{r=1}^{P}e^{-r/s}
},
\qquad
\beta_s^-(\tau)
=
\frac{
\mathbf{1}[\tau<0]e^{\tau/s}
}{
\sum_{r=1}^{P}e^{-r/s}
},
\end{equation}
for exponentially spaced scales
\[
s\in\{1,2,4,\ldots\le P\}.
\]
In the reported experiments, $P$ is set to the dataset seasonal length.

The response kernel is parameterized as
\[
g_{k,m,f}(\tau)
=
\sum_{j=1}^{J}
W_{k,m,j,f}\beta_j(\tau),
\]
where $k$, $m$, $j$, and $f$ index event type, relative-position component, temporal basis, and target variable, respectively.

To allow forecasting-unit heterogeneity while retaining a shared response structure, Loki applies a positive low-rank modulation,
\begin{equation}
\label{eq:app-entity-modulation}
a_{i,k,m}
=
\exp
\left(
\operatorname{clip}_{[-2,2]}
\left(\langle V_i,U_{k,m}\rangle\right)
\right).
\end{equation}
where $V_i\in\mathbb{R}^{R}$ is a learnable forecasting-unit embedding and $U_{k,m}\in\mathbb{R}^{R}$ is a learnable vector for event type $k$ and representation channel $m$, shared across forecasting units.

The response associated with event type $k$ is then
\[
s_{i,t,k,f}
=
\sum_m
a_{i,k,m}
\left(
g_{k,m,f} * \psi_{i,k,m}
\right)_t,
\]
and the total log response is obtained by summing across event types,
\[
S_{i,t,f}
=
\sum_k s_{i,t,k,f}.
\]

Thus, event types share the same response formulation while retaining event-specific, target-specific, and forecasting-unit-specific structure.

\subsection{Cross-Boundary Computation}
\label{app:cross-boundary}

The complete history-and-forecast interval is treated as a single response domain. The ERF is evaluated once on the full event schedule and only then partitioned into
\[
S^{\mathrm{hist}}
=
S_{1:L},
\qquad
S^{\mathrm{fut}}
=
S_{L+1:L+H}.
\]

The forecasting boundary is therefore \emph{not} a response boundary: a historical event may have a persistent effect extending into the forecast horizon, while a future event may produce an anticipatory response extending into the historical window. Only the external boundaries of the complete $L+H$ window truncate the response. \textcolor{black}{Appendix~\ref{app:future-events-history} evaluates removing future new events from the historical response.}

For efficiency, the bidirectional response is implemented using zero-padded FFT-based linear convolution.

Lift Out and Kick In are both applied in the original target space.
\[
\textcolor{black}{X^0 = X \odot \exp\!\left(-S^{\mathrm{hist}}\right),}
\]
and
\[
\textcolor{black}{\widehat Y = \widehat Y^0 \odot \exp\!\left(S^{\mathrm{fut}}\right).}
\]

Window-level RevIN is applied only between these two operations: after Lift Out and before forecasting, and inverted before Kick In.

\subsection{Forecasting and Learning Details}
\label{app:forecasting-learning}

The default temporal forecaster follows a lightweight SparseTSF-style design and is augmented with a recent-state branch that encodes the most recent seasonal segment. The two forecasting branches are combined additively.

For datasets with environmental covariates, Loki uses their future deviation from the historical mean,
\begin{equation}
\label{eq:app-context-deviation}
\Delta C_t
=
C_t
-
\frac{1}{L}
\sum_{\ell=1}^{L}C_\ell,
\end{equation}
where $t$ indexes a forecast position. Let $\chi_t$ denote the corresponding calendar/time representation and let $z_t$ denote the backbone latent state at that position. A lightweight network $h_{\phi}$ produces the FiLM parameters through zero-reference subtraction,
\begin{equation}
\label{eq:app-film-parameters}
[\lambda_t,\delta_t]
=
h_{\phi}([\Delta C_t,\chi_t])-h_{\phi}([0,\chi_t]),
\end{equation}
\begin{equation}
\label{eq:app-film-modulation}
z_t'
=
(1+\lambda_t)\odot z_t+\delta_t.
\end{equation}

The reference subtraction ensures zero modulation when the future environment matches the historical reference state.

Loki is trained only through the final scale-normalized forecasting loss,
\[
\mathcal L
=
\frac{1}{BHF}
\sum_{b=1}^{B}
\sum_{h=1}^{H}
\sum_{f=1}^{F}
\left|
\frac{
\widehat Y_{b,h,f}
-
Y_{b,h,f}
}{
\sigma_f^{\mathrm{train}}
}
\right|,
\]
with no direct supervision on the event response or decomposition components.

\textcolor{black}{The final forecasting loss jointly updates the backbone and the shared ERF through both historical adjustment and future response generation.}

\begin{table}[H]
\centering
\color{black}
\captionsetup{labelfont={color=black},textfont={color=black}}
\caption{Gradient paths from the forecasting loss. Arrows follow backpropagation to the backbone parameters $\theta$ or the shared ERF parameters $\omega$.}
\label{tab:forecast-gradient-paths}
\begin{tabular}{@{}ll@{}}
\toprule
Path & Backpropagation order \\
\midrule
Kick In  & $\mathcal L \longrightarrow \widehat Y \longrightarrow S^{\mathrm{fut}} \longrightarrow \omega$ \\
Backbone & $\mathcal L \longrightarrow \widehat Y \longrightarrow \widehat Y^0 \longrightarrow \theta$ \\
Lift Out & $\mathcal L \longrightarrow \widehat Y \longrightarrow \widehat Y^0 \xrightarrow{B_\theta} X^0 \longrightarrow S^{\mathrm{hist}} \longrightarrow \omega$ \\
\bottomrule
\end{tabular}
\end{table}

Finally, the ERF is initialized to produce the identity multiplicative effect. The temporal response weights are initialized to zero, yielding $S=0$ and hence $\exp(S)=1$; forecasting-unit modulation is initialized to one; the event-intensity power is initialized to $\gamma_k=1$; and the environmental FiLM branch is initialized to zero modulation.

% \subsection{Implementation Configuration}
% \label{app:implementation-config}

% Table~\ref{tab:loki-config} summarizes the key implementation choices used in the reported experiments, including the temporal scales and response support of the ERF, forecasting-unit modulation, response clipping, and initialization. Unless otherwise specified, these configurations are shared across datasets; dataset-dependent quantities are derived from the corresponding temporal resolution and seasonal length.

% \begin{table}[t]
% \centering
% \caption{Key implementation configurations of Loki.}
% \label{tab:loki-config}
% \small
% \begin{tabular}{ll}
% \toprule
% Component & Configuration \\
% \midrule
% Position scales
% & steps-per-day, seasonal length \\
% Position components
% & sustained, start proximity, end proximity \\
% Response support
% & $P=$ dataset seasonal length \\
% Temporal basis scales
% & $1,2,4,\ldots\le P$ \\
% Forecasting-unit modulation rank
% & $R=8$ \\
% Response clipping
% & $[-8,8]$ \\
% Event-intensity initialization
% & $\gamma_k=1$ \\
% ERF initialization
% & $W=0$ \\
% Environmental modulation
% & FiLM with zero-reference subtraction \\
% ERF supervision
% & final forecasting loss only \\
% \bottomrule
% \end{tabular}
% \end{table}

\section{Dataset and Forecasting Protocol Details}
\label{app:data-protocol}

\subsection{Forecasting Protocol}
\label{app:forecasting-protocol}

We construct rolling-origin samples under fixed chronological splits. Forecast horizons lie entirely within one split, while historical inputs may precede its boundary. Table~\ref{tab:dataset-overview} summarizes the statistics of the five benchmark datasets. Table~\ref{tab:forecasting-protocol-app} summarizes the window configurations and sample counts.

\begin{table}[t]
\input{table/benchmark}
\par
\vspace{10pt}
\centering
\captionsetup{type=table}
\caption{Forecasting protocol across datasets. History and forecast lengths are reported in time steps. Sample counts are forecasting windows after split and continuity filtering.}
\label{tab:forecasting-protocol-app}
\begingroup
\footnotesize
\setlength{\tabcolsep}{3pt}
\begin{tabular}{lcccc}
\toprule
Dataset & History $L$ & Forecast $H$ & Stride & \makecell{Samples\\Train / Val / Test} \\
\midrule
Ride-hailing
& 1344 & 336 & 48
& 212,200 / 35,200 / 35,600 \\

TaxiBJ
& 336 & 48 & 48
& 150,528 / 29,696 / 20,480 \\

M5
& 364 & 28 & 1
& 106,539 / 70 / 70 \\

Rossmann
& 364 & 28 & 1
& 275,599 / 62,645 / 151,470 \\

Seoul Bike
& 168 & 24 & 24
& 248 / 36 / 73 \\
\bottomrule
\end{tabular}
\endgroup
\end{table}

TaxiBJ and Rossmann contain temporal discontinuities. We require the complete $L+H$ span of each sample to be temporally contiguous and discard any window crossing a gap. No temporal padding or missing-timestamp interpolation is used.

We evaluate future-exogenous models under a \textbf{known-future covariate} setting: exogenous variables supplied by each benchmark protocol are treated as available throughout the forecast horizon, whereas future target values are never provided as model inputs.

\textcolor{black}{Seoul Bike comprises one year of hourly observations from a single entity, with 255.5 days used for training. The training, validation and test partitions contain 8, 1 and 3 contiguous holiday intervals, respectively. All zero targets correspond to non-operating periods, whose proportion increases from 0.78\% in training to 12.73\% in testing. Together with the limited holiday coverage, this operating-status shift makes it more difficult to evaluate event effects.}

\subsection{Covariates and Dataset Construction}
\label{app:covariate-construction}

Table~\ref{tab:covariate-roles-app} summarizes the dataset- specific assignment of covariates to modeling roles based on semantics rather than a universal variable-type rule. For ERF inputs, categorical events are expanded into event-type channels, while binary and continuous values encode activity and intensity, respectively.
% Loki assigns covariates to different modeling roles according to their dataset-specific semantics. Event channels are modeled by the ERF, continuously varying contextual variables condition the forecasting backbone, and additional calendar, entity, or operating-status information is used as auxiliary input where applicable. Table~\ref{tab:covariate-roles-app} summarizes the resulting assignment.

\begin{table}[t]
\centering
\caption{High-level covariate roles used by Loki across datasets.}
\label{tab:covariate-roles-app}
\footnotesize
\begin{tabular}{llll}
\toprule
Dataset & ERF channels & Backbone context & Other known inputs \\
\midrule
Ride-hailing
& event and calendar signals
& --
& calendar, entity \\

TaxiBJ
& holidays
& temperature, wind, weather
& calendar, grid \\

M5
& SNAP, calendar events
& price, promotion aggregates
& calendar, entity \\

Rossmann
& promotion, holidays
& \textcolor{black}{open status}
& \textcolor{black}{calendar, store} \\

Seoul Bike
& holiday
& \textcolor{black}{weather, operating status}
& \textcolor{black}{calendar} \\
\bottomrule
\end{tabular}
\end{table}

% The event/context assignment is dataset-specific rather than inferred from a universal variable-type rule. Categorical events are expanded into event-type channels, while binary or continuous event values retain their activity or intensity information before entering the ERF.

We retain only preprocessing steps that change the forecasting problem definition. TaxiBJ windows crossing temporal gaps are removed. M5 is evaluated at the store--department level using Level-9 aggregated sales, with corresponding price, promotion, SNAP, and calendar covariates. For Rossmann, unobserved sales records are excluded while closed-store periods are retained; same-day customer counts are not used as model inputs. Seoul Bike retains non-functioning periods and \textcolor{black}{provides operating status as an environmental covariate to the backbone}.

\begingroup
\color{black}
\paragraph{Input information across models.}
Table~\ref{tab:covariate-sources} lists the event and environmental variables, and Table~\ref{tab:model-input-scope} specifies their historical/forecast usage and the calendar information supplied to each model. Target values are not included in these covariate lists. Models with known-future covariates use common event and environmental sources, while retaining their own encodings and any additional calendar information. Loki receives no exclusive event or environmental observations.

\begin{table}[!htbp]
\centering
\color{black}
\caption{\textcolor{black}{Event and environmental covariates across the five datasets. Operating status is included in environmental information.}}
\label{tab:covariate-sources}
\scriptsize
\setlength{\tabcolsep}{4pt}
\renewcommand{\arraystretch}{1.15}
\begin{tabular}{@{}>{\raggedright\arraybackslash}p{0.14\linewidth}>{\raggedright\arraybackslash}p{0.45\linewidth}>{\raggedright\arraybackslash}p{\dimexpr0.41\linewidth-16pt\relax}@{}}
\toprule
Dataset & Event information & Environmental information \\
\midrule
Ride-Hailing & Weather factor, two types of social activity, public holidays, traditional festivals, Western festivals & No separately designated environmental variables \\
\midrule[0.3pt]
TaxiBJ & Holiday type & Temperature, wind speed, weather category (16 categories) \\
\midrule[0.3pt]
M5 & SNAP food-assistance indicators, including state-specific indicators; cultural, national, religious and sporting events; a second same-day event (cultural or religious) & Price index, promotion share \\
\midrule[0.3pt]
Rossmann & Promotions, public holidays, school holidays & Store operating status \\
\midrule[0.3pt]
Seoul Bike & Holidays & Temperature, humidity, wind speed, visibility, dew-point temperature, solar radiation, rainfall, snowfall, and service operating status \\
\bottomrule
\end{tabular}
\end{table}

\begin{table}[!htbp]
\centering
\color{black}
\caption{\textcolor{black}{Input scope and calendar information by model. Event/environment scope applies to the variables in Table~\ref{tab:covariate-sources} across all five datasets, subject to the filtering and encoding differences below. Calendar inputs follow the same historical/forecast scope as the corresponding row.}}
\label{tab:model-input-scope}
\scriptsize
\setlength{\tabcolsep}{3pt}
\renewcommand{\arraystretch}{1.12}
\begin{tabular}{@{}>{\raggedright\arraybackslash}p{0.14\linewidth}>{\raggedright\arraybackslash}p{0.15\linewidth}>{\raggedright\arraybackslash}p{0.23\linewidth}>{\raggedright\arraybackslash}p{0.23\linewidth}>{\raggedright\arraybackslash}p{\dimexpr0.25\linewidth-24pt\relax}@{}}
\toprule
Model & Event/\allowbreak environment scope & Ride-Hailing calendar & TaxiBJ / Seoul Bike calendar & M5 / Rossmann calendar \\
\midrule
DLinear & Not used & No explicit input & No explicit input & No explicit input \\
PatchTST & Not used & No explicit input & No explicit input & No explicit input \\
SparseTSF & Not used & No explicit input & No explicit input & No explicit input \\
TimeMixer & Not used & No explicit input & No explicit input & No explicit input \\
iTransformer & Not used & No explicit input & No explicit input & No explicit input \\
\midrule
TimeXer & History only & Day of week, time of day, week of year & No explicit input & No explicit input \\
CrossLinear & History only & Day of week, time of day, week of year & No explicit input & No explicit input \\
XLinear & History only & Day of week, time of day, week of year & No explicit input & No explicit input \\
\midrule
TiDE & History + forecast & Day of week, time of day, week of year & Day of week, time of day & Day of week \\
DAG & History + forecast & Day of week, time of day, week of year & Day of week, time of day & Day of week \\
TFT & History + forecast & Day of week, time of day, week of year & Day of week, time of day & Day of week \\
AutoARIMA-X\textsuperscript{1} & History + forecast & Within-day, within-week and within-year position & Within-day, within-week and within-year position & Within-week and within-year position \\
Log-linear\textsuperscript{2} & History + forecast & Within-day, within-week and within-year position & Within-day, within-week and within-year position & Within-week and within-year position \\
Prophet\textsuperscript{3} & History + forecast & Day of week, time of day, week of year & Day of week, time of day & Day of week \\
NeuralProphet & History + forecast & Day of week, time of day, within-year position & Day of week, time of day, within-year position & Day of week, within-year position \\
NBEATSx & History + forecast & Day of week, time of day, week of year & Day of week, time of day & Day of week \\
SPADE & History + forecast & Day of week, time of day, week of year & Day of week, time of day & Day of week \\
CEDAR & History + forecast & Day of week, time of day & Day of week, time of day & Day of week \\
\textbf{Loki} & \textbf{History + forecast} & \textbf{Day of week, time of day} & \textbf{Day of week, time of day} & \textbf{Day of week} \\

\bottomrule
\end{tabular}
\par\smallskip
\begin{minipage}{\linewidth}
\scriptsize\color{black}
\textsuperscript{1}AutoARIMA-X removes constant and linearly dependent columns during fitting; retained variables can differ across entities.
\textsuperscript{2}Log-linear assigns zero coefficients to covariates that remain constant during fitting.
\textsuperscript{3}Prophet constructs regressors from categories observed during training; unseen categories receive no separate terms.
\end{minipage}
\end{table}

``Forecast'' denotes covariates supplied for the forecast window in these experiments. Weather inputs use actual future observations from the datasets.

\endgroup
\FloatBarrier

\section{Experimental Implementation Details}
\label{app:experimental-implementation}

\subsection{Training and Model Selection}
\label{app:optimization-config}

We train Loki in PyTorch with AdamW using Python 3.11. The GPU experiments run on NVIDIA RTX 6000D GPUs. \textcolor{black}{For models trained with early stopping,} we evaluate standardized validation MAE after each epoch and select the checkpoint with the lowest value. Patience is model-specific. DAG includes an auxiliary causality loss with coefficient $\beta=0.1$. \textcolor{black}{Table~\ref{tab:baseline-config} describes the fitting procedures for AutoARIMA-X, log-linear regression, and Prophet.}

\paragraph{\textcolor{black}{Event Lookup.}}
\textcolor{black}{Event Lookup learns a response coefficient for each event type, temporal offset, and target channel. It multiplies event intensities at offsets in $[-P,P]$ by the corresponding coefficients and sums the contributions to obtain the log response. The table is zero-initialized and shared across entities and across historical and forecast windows. The backbone, Lift Out/Kick In operations, event inputs, response support, and training configuration are unchanged from Loki.}

\paragraph{\textcolor{black}{Model Size and Computational Cost.}}
\begingroup
\color{black}
We measure neural-model costs on Ride-Hailing using the same NVIDIA RTX 6000D GPU. Training time is the measured per-epoch time multiplied by the number of epochs actually run. Inference covers the complete test set and is averaged per forecast window. AutoARIMA-X, log-linear regression, and Prophet are fitted per series on CPU and are excluded from Table~\ref{tab:model-cost}.

\begin{table}[!htbp]
\centering
\small
\color{black}
\caption{\textcolor{black}{Model size and computational cost on Ride-Hailing. M denotes one million parameters.}}
\label{tab:model-cost}
\begin{tabular}{lrrr}
\toprule
Model & Trainable parameters & \shortstack{Training time\\(min)} & \shortstack{Inference time\\(ms/window)} \\
\midrule
DLinear & 0.90M & 2.3 & 0.07 \\
PatchTST & 1.66M & 11.6 & 0.07 \\
SparseTSF & 245 & 3.1 & 0.07 \\
TimeMixer & 10.97M & 30.1 & 0.38 \\
iTransformer & 0.48M & 5.8 & 0.06 \\
TimeXer & 3.02M & 24.8 & 0.14 \\
CrossLinear & 2.01M & 3.5 & 0.08 \\
XLinear & 3.12M & 2.5 & 0.07 \\
TiDE & 5.45M & 8.5 & 0.06 \\
DAG & 17.21M & 46.9 & 0.31 \\
TFT & 0.95M & 175.8 & 0.73 \\
NBEATSx & 1.98M & 10.5 & 0.08 \\
NeuralProphet & 0.11M & 6.9 & 0.07 \\
SPADE & 0.23M & 110.1 & 0.40 \\
CEDAR & 4.78M & 90.3 & 15.83 \\
Loki & 0.27M & 18.1 & 0.20 \\
\bottomrule
\end{tabular}
\end{table}

Loki has 0.27M trainable parameters, fewer than all neural baselines except SparseTSF, NeuralProphet, and SPADE.
\endgroup
\FloatBarrier

\subsection{Baseline Implementations and Configurations}
\label{app:baseline-config}

Table~\ref{tab:baseline-config} lists the implementation sources and key configurations of all 18 baselines in Table~\ref{tab:overall-performance}. Official model implementations are connected to the benchmark through data and training adapters where needed. SPADE and CEDAR have no public code; we reimplement them following their papers \citep{wolff2024spade,meng2026cedar}, with the adaptations specified in the table.

\begin{table}[!htb]
\centering
\caption{Baseline implementation sources and key model configurations.}
\label{tab:baseline-config}
\scriptsize
\setlength{\tabcolsep}{4pt}
\renewcommand{\arraystretch}{1.15}
\begin{tabular}{@{}>{\raggedright\arraybackslash}p{\dimexpr.17\textwidth-8pt\relax}>{\raggedright\arraybackslash}p{\dimexpr.25\textwidth-8pt\relax}>{\raggedright\arraybackslash}p{\dimexpr.58\textwidth-8pt\relax}@{}}
\toprule
Model & Implementation source & Key configuration \\
\midrule
DLinear
& Official implementation
& individual = false \\
\addlinespace[2pt]

PatchTST
& Official implementation
& patch = 48, stride = 48, $d_{\mathrm{model}}=128$, 8 heads, 3 layers \\
\addlinespace[2pt]

SparseTSF
& Official implementation
& linear variant; period set according to dataset periodicity \\
\addlinespace[2pt]

TimeMixer
& Official implementation
& $d_{\mathrm{model}}=32$, $d_{\mathrm{ff}}=64$, 2 layers, 2 downsampling layers \\
\addlinespace[2pt]

iTransformer
& Official implementation
& $d_{\mathrm{model}}=128$, 8 heads, 2 layers \\
\addlinespace[2pt]

TimeXer
& Official implementation
& $d_{\mathrm{model}}=128$, 8 heads, 2 layers; dataset-dependent patch length \\
\addlinespace[2pt]

CrossLinear
& Official implementation
& patch = 24, $d_{\mathrm{model}}=128$, $d_{\mathrm{ff}}=256$ \\
\addlinespace[2pt]

XLinear
& Official implementation
& $d_{\mathrm{model}}=1024$, temporal FF = 256 \\
\addlinespace[2pt]

TiDE
& PyTorch reimplementation
& hidden = 256, 2 layers, series embedding = 16 \\
\addlinespace[2pt]

DAG
& Official implementation
& patch = 48, stride = 48, $d_{\mathrm{model}}=64$, 4 heads, 1 layer \\
\addlinespace[2pt]

TFT
& PyTorch adaptation
& hidden = 64, 4 heads, dropout = 0.1 \\
\addlinespace[2pt]

AutoARIMA-X
& StatsForecast AutoARIMA + adapter
& ARIMA errors with Fourier and exogenous regressors; fit on training data and reuse fitted parameters for each forecast window. \\
\addlinespace[2pt]

Log-linear (SCAN*PRO)
& NumPy ridge regression (SVD)
& Regress the shifted log target on Fourier and exogenous terms; choose ridge penalty by training GCV; adjust the forecast level using recent log residuals. \\
\addlinespace[2pt]

Prophet
& Official package + rolling-fit adapter
& Refit on the current history window at each forecast origin; weekly seasonality enabled; deterministic MAP estimation. \\
\addlinespace[2pt]

NeuralProphet
& Official TimeNet + training adapter
& AR hidden size = 64; yearly/weekly Fourier orders = 6/3, daily order = 6 for subdaily data; share TimeNet across entities and target channels. \\
\addlinespace[2pt]

NBEATSx
& Official model core + training adapter
& Trend, seasonal, and exogenous stacks; 3 weight-sharing blocks per stack; 2-layer MLP with 256 hidden units per layer. \\
\addlinespace[2pt]

SPADE
& Independent paper-based reimplementation
& MQCNN-style backbone without MQT decoder attention; forward-fill historical peaks and predict future peaks; P50/P90 pinball loss, with P50 as the point forecast. \\
\addlinespace[2pt]

CEDAR
& Independent paper-based reimplementation
& Two-stage AIT/RC training; freeze AIT in stage two; replace text embeddings with structured event inputs and entity embeddings. \\
\bottomrule
\end{tabular}
\end{table}

For TimeXer, the patch length is 24 for Ride-hailing, TaxiBJ, and Seoul Bike, and 7 for M5 and Rossmann.

\section{Additional Experimental Results}
\label{app:additional-results}

\subsection{Additional Evaluation Metrics}
\label{app:additional-metrics}

In addition to the standardized MAE and MSE reported in the main paper, we report standardized RMSE and original-scale WMAPE as complementary evaluation metrics. Tables~\ref{tab:appendix-rmse} and~\ref{tab:appendix-wmape} report the results. For each random seed, standardized RMSE is computed as

\[
\mathrm{RMSE}_{\mathrm{std}}
=
\sqrt{\mathrm{MSE}_{\mathrm{std}}},
\]

and is then averaged across seeds. Original-scale WMAPE is computed independently for each target channel as

\begin{equation}
\label{eq:app-wmape}
\mathrm{WMAPE}_c
=
\frac{
\sum_{n,h}
\left|
\widehat{y}_{n,h,c}-y_{n,h,c}
\right|
}{
\sum_{n,h}
\left|y_{n,h,c}\right|
},
\end{equation}

followed by an equal average across target channels with nonzero denominators. Predictions and targets are evaluated in the original target scale. Reported uncertainties are sample standard deviations across seeds. \textcolor{black}{The tables use the same 19 models and evaluated runs as Table~\ref{tab:overall-performance}; deterministic fits are reported without a seed standard deviation.}

\input{table/appendix_rmse}

\input{table/appendix_wmape}

Loki achieves the lowest original-scale WMAPE on all five datasets, consistent with the standardized MAE results reported in the main paper. \textcolor{black}{It also achieves the lowest standardized RMSE on Ride-Hailing, TaxiBJ, M5, and Rossmann, and the second-lowest on Seoul Bike.}

\FloatBarrier

\subsection{\textcolor{black}{Parameter-Budget-Matched Independent ERF}}
\label{app:capacity-ablation}
\begingroup
\color{black}
We compare the shared ERF, two full-capacity independent ERFs, and Half-independent while keeping the forecasting backbone fixed. Half-independent reduces the entity-modulation rank and adjusts the number of temporal bases to approximately match the shared ERF's total parameter budget. The rank and number of temporal-basis pairs per ERF are (6, 4), (4, 4), (4, 1), (4, 1), and (2, 5) for Ride-Hailing, TaxiBJ, M5, Rossmann, and Seoul Bike, respectively. Within each dataset, all three variants use the same training configuration, including backbone dropout of 0.1 and zero weight decay. Table~\ref{tab:capacity-ablation} reports the combined ERF parameter count and prediction errors across multiple seeds.

\begin{table}[htbp]
\centering
\small
\color{black}
\caption{\textcolor{black}{Shared and independent ERFs with the same forecasting backbone. Parameter counts include the ERF components only; prediction MAE and MSE are means $\pm$ sample standard deviations across seeds.}}
\label{tab:capacity-ablation}
\begin{adjustbox}{max width=\linewidth}
\begin{tabular}{llrcc}
\toprule
Dataset & Variant & ERF parameters & Prediction MAE & Prediction MSE \\
\midrule
\multirow{3}{*}{Ride-Hailing} & Shared & 32,960 & $0.15696\pm0.00031$ & $0.19786\pm0.00057$ \\
 & Independent & 65,920 & $0.15717\pm0.00044$ & $0.19745\pm0.00137$ \\
 & Half-independent & 32,920 & $0.15691\pm0.00019$ & $0.19809\pm0.00084$ \\
\midrule
\multirow{3}{*}{TaxiBJ} & Shared & 9,809 & $0.11276\pm0.00161$ & $0.04749\pm0.00119$ \\
 & Independent & 19,618 & $0.11885\pm0.00321$ & $0.05244\pm0.00260$ \\
 & Half-independent & 9,746 & $0.11958\pm0.00201$ & $0.05314\pm0.00187$ \\
\midrule
\multirow{3}{*}{M5} & Shared & 1,320 & $0.09583\pm0.00109$ & $0.02502\pm0.00066$ \\
 & Independent & 2,640 & $0.09825\pm0.00249$ & $0.02756\pm0.00236$ \\
 & Half-independent & 1,280 & $0.09845\pm0.00136$ & $0.02722\pm0.00091$ \\
\midrule
\multirow{3}{*}{Rossmann} & Shared & 9,300 & $0.13649\pm0.00280$ & $0.04741\pm0.00199$ \\
 & Independent & 18,600 & $0.13963\pm0.00208$ & $0.05022\pm0.00116$ \\
 & Half-independent & 9,280 & $0.14631\pm0.00224$ & $0.05381\pm0.00111$ \\
\midrule
\multirow{3}{*}{Seoul Bike} & Shared & 134 & $0.32379\pm0.02080$ & $0.24733\pm0.02312$ \\
 & Independent & 268 & $0.32234\pm0.02474$ & $0.24345\pm0.02786$ \\
 & Half-independent & 136 & $0.30801\pm0.00910$ & $0.22638\pm0.00658$ \\
\bottomrule
\end{tabular}
\end{adjustbox}
\end{table}

Sharing gives lower mean prediction MAE and MSE than both independent variants on TaxiBJ, M5, and Rossmann. On Ride-Hailing, differences are small relative to the variation across seeds. On Seoul Bike, Half-independent has the lowest mean MAE and MSE among the three variants. Reducing independent ERF capacity therefore does not have a uniform effect across datasets. Because Half-independent also changes rank and temporal bases, this control does not isolate parameter sharing from all differences in representation capacity. See Appendix~\ref{app:forecasting-protocol} for Seoul Bike's event coverage and operating-status distribution.
\endgroup

\subsection{\textcolor{black}{Additive and Multiplicative Decomposition}}
\label{app:additive-ablation}
\begingroup
\color{black}
We compare additive and multiplicative decompositions using the same forecasting backbone, shared ERF, event inputs, and training protocol. The additive variant subtracts the estimated event increment from history and adds the future increment to the backbone forecast; the ERF output is scaled by the target's training standard deviation to express increments in target units. Both variants are trained jointly through the final forecasting loss. Table~\ref{tab:additive-ablation} reports prediction MAE and MSE on the original real-data benchmarks.

\begin{table}[htbp]
\centering
\small
\color{black}
\caption{\textcolor{black}{Prediction MAE and MSE for multiplicative and additive structures with the same backbone and shared ERF. Values are means $\pm$ sample standard deviations across training seeds.}}
\label{tab:additive-ablation}
\begin{adjustbox}{max width=\textwidth}
\begin{tabular}{lcccc}
\toprule
\multirow{2}{*}{Dataset} & \multicolumn{2}{c}{Prediction MAE} & \multicolumn{2}{c}{Prediction MSE} \\
\cmidrule(lr){2-3}\cmidrule(lr){4-5}
 & Multiplicative structure & Additive structure & Multiplicative structure & Additive structure \\
\midrule
Ride-Hailing & $0.15696\pm0.00031$ & $0.15791\pm0.00017$ & $0.19786\pm0.00057$ & $0.20346\pm0.00037$ \\
TaxiBJ & $0.11276\pm0.00161$ & $0.13127\pm0.02151$ & $0.04749\pm0.00119$ & $0.06678\pm0.02432$ \\
M5 & $0.09583\pm0.00109$ & $0.09663\pm0.00133$ & $0.02502\pm0.00066$ & $0.02533\pm0.00118$ \\
Rossmann & $0.13649\pm0.00280$ & $0.15898\pm0.00349$ & $0.04741\pm0.00199$ & $0.06081\pm0.00234$ \\
Seoul Bike & $0.32379\pm0.02080$ & $0.33622\pm0.03582$ & $0.24733\pm0.02312$ & $0.26190\pm0.04033$ \\
\bottomrule
\end{tabular}
\end{adjustbox}
\end{table}

The multiplicative formulation has lower mean prediction MAE and MSE on all five datasets. This comparison supports its use in these tasks; the semi-synthetic comparisons in Appendix~\ref{app:controlled-recovery} separately examine recovery when the injected effect does or does not match the model's response form.
\endgroup

\subsection{Event-Regime Analysis}
\label{app:event-regime-analysis}

We further analyze forecasting performance according to whether active discrete-event indicators appear in the historical and forecast windows. Each test window is assigned to one of four regimes: Quiet$\rightarrow$Quiet (Q$\rightarrow$Q), Event$\rightarrow$Quiet (E$\rightarrow$Q), Quiet$\rightarrow$Event (Q$\rightarrow$E), or Event$\rightarrow$Event (E$\rightarrow$E). Table~\ref{tab:appendix-event-regime} reports the corresponding standardized MAE. \textcolor{black}{The analysis uses the same five seeds as Table~\ref{tab:ablation}. Weighting the four regime MAEs by their sample counts recovers the overall Ride-Hailing MAE for each variant.}

\input{table/appendix_event_regime}

Full Loki has the lowest mean MAE in Q$\rightarrow$Q, E$\rightarrow$Q, and E$\rightarrow$E. Its largest gain over the Backbone occurs in E$\rightarrow$E (\textcolor{black}{6.25\%}). \textcolor{black}{In Q$\rightarrow$E, Future-only has the lowest mean MAE. The relative MAE reductions of Full Loki over Future-only are $-0.53$\%, $+0.26$\%, $-0.60$\%, $-0.48$\%, and $-0.36$\% for seeds 2027--2031, respectively ($-0.34$\% using the mean MAEs across seeds). Of the 1,269 windows in this regime, 1,127 contain a traditional festival as their only discrete event type in the forecast horizon. On this subset, the paired MAE differences (Full Loki minus Future-only) are $+0.00104$, $-0.00012$, $+0.00109$, $+0.00101$, and $+0.00072$ for seeds 2027--2031, respectively. On the 129 windows containing social events alone, Full Loki has lower mean MAE in all five runs.}

\FloatBarrier
\Needspace{6\baselineskip}
\subsection{\textcolor{black}{Sensitivity to Temporal Scales}}
\label{app:temporal-scale-sensitivity}
\begingroup
\color{black}
We vary the backbone period or ERF support radius while keeping other training settings unchanged. Table~\ref{tab:temporal-scale-sensitivity} reports multi-seed mean standardized MAE and percentage changes relative to the default. Event regions are fixed across configurations.

\begin{table}[htbp]
\centering
\scriptsize
\color{black}
\setlength{\tabcolsep}{3pt}
\caption{\textcolor{black}{Temporal-scale sensitivity. Parentheses show the MAE change relative to the default configuration. Event-active positions are defined by the event indicators; inactive positions before and after events use the default seven-day support across all configurations.}}
\label{tab:temporal-scale-sensitivity}
\begin{adjustbox}{max width=\textwidth}
\begin{tabular}{llccccc}
\toprule
\multirow{2}{*}{Dataset} & \multirow{2}{*}{Configuration} & \multirow{2}{*}{Period} & \multirow{2}{*}{Support radius} & \multicolumn{3}{c}{Prediction MAE} \\
\cmidrule(lr){5-7}
 & & & & Overall & Event-active & Before/after events (inactive) \\
\midrule
\multirow{6}{*}{Ride-Hailing} & Default & 24 h & 7 d & 0.15696 (0.00\%) & 0.20543 (0.00\%) & 0.13914 (0.00\%) \\
 & Shorter period & 12 h & 7 d & 0.15722 (+0.17\%) & 0.20563 (+0.10\%) & 0.13943 (+0.21\%) \\
 & Longer period & 48 h & 7 d & 0.15810 (+0.73\%) & 0.20620 (+0.38\%) & 0.14041 (+0.91\%) \\
 & Nonaligned period & 16 h & 7 d & 0.15909 (+1.36\%) & 0.20674 (+0.64\%) & 0.14156 (+1.74\%) \\
 & Shorter support & 24 h & 3.5 d & 0.15687 (-0.06\%) & 0.20594 (+0.25\%) & 0.13883 (-0.22\%) \\
 & Longer support & 24 h & 14 d & 0.15681 (-0.09\%) & 0.20510 (-0.16\%) & 0.13905 (-0.07\%) \\
\midrule
\multirow{6}{*}{M5} & Default & 7 d & 7 d & 0.09583 (0.00\%) & 0.09978 (0.00\%) & 0.08973 (0.00\%) \\
 & Shorter period & 1 d & 7 d & 0.09425 (-1.65\%) & 0.09820 (-1.58\%) & 0.08814 (-1.78\%) \\
 & Longer period & 14 d & 7 d & 0.09709 (+1.32\%) & 0.10268 (+2.91\%) & 0.08845 (-1.42\%) \\
 & Nonaligned period & 4 d & 7 d & 0.09697 (+1.18\%) & 0.10130 (+1.52\%) & 0.09027 (+0.60\%) \\
 & Shorter support & 7 d & 4 d & 0.09668 (+0.88\%) & 0.10013 (+0.35\%) & 0.09133 (+1.79\%) \\
 & Longer support & 7 d & 14 d & 0.09812 (+2.39\%) & 0.10482 (+5.05\%) & 0.08777 (-2.19\%) \\
\bottomrule
\end{tabular}
\end{adjustbox}
\end{table}

Loki shows some tolerance to the tested changes in period and response range, with overall MAE increasing by at most 1.36\% on Ride-Hailing and 2.39\% on M5. Both nonaligned and longer periods increase overall error, indicating that period selection still matters and that longer periods are not necessarily better. Changes can differ between event-active and surrounding inactive regions: for example, wider response support on M5 increases MAE in the former by 5.05\% but reduces it in the latter by 2.19\%, so overall changes can mask regional differences. This study focuses on time series with an exploitable periodic background; these results do not establish robustness to continuously drifting or irregular periodicity.
\endgroup
\FloatBarrier

\Needspace{6\baselineskip}
\subsection{\textcolor{black}{Forecast Distance and Temporal Consistency}}
\label{app:forecast-distance-consistency}
\begingroup
\color{black}
We evaluate errors at each forecast step and errors in changes between adjacent steps using the saved test predictions. For each dataset, we compare Loki with the baseline having the lowest overall MAE in the main evaluation, keeping this comparator fixed across forecast steps and both analyses. Results are averaged across training seeds.

\paragraph{Forecast distance.}
Figure~\ref{fig:appendix-horizon-steps} shows standardized MAE at each forecast step. Table~\ref{tab:appendix-horizon-halves} summarizes the first and second halves of the forecast window. Loki achieves the lowest first-half MAE on all five datasets and retains the lowest second-half MAE on four of them.

\begin{figure}[htbp]
\centering
\includegraphics[width=\linewidth]{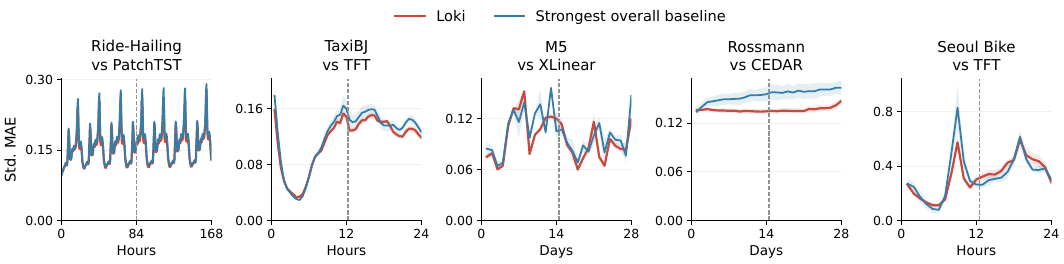}
\caption{\textcolor{black}{Prediction error across forecast distances. Curves show the multi-seed mean standardized MAE at each forecast step; shading indicates $\pm1$ standard deviation. Each dataset uses its best-overall baseline throughout. Vertical dashed lines divide the forecast window into its first and second halves.}}
\label{fig:appendix-horizon-steps}
\end{figure}

\begin{table}[htbp]
\centering
\small
\color{black}
\caption{\textcolor{black}{Prediction MAE in the first and second halves of the forecast window. Both MAE and ranks follow the order first half / second half. Baselines are the same as in Figure~\ref{fig:appendix-horizon-steps}; Loki's ranks are computed over all 19 models.}}
\label{tab:appendix-horizon-halves}
\setlength{\tabcolsep}{4pt}
\begin{adjustbox}{max width=\linewidth}
\begin{tabular}{llccc}
\toprule
Dataset & Best-overall baseline & Loki MAE & Baseline MAE & Loki rank \\
\midrule
Ride-Hailing & PatchTST & 0.1507 / 0.1632 & 0.1545 / 0.1690 & 1 / 1 \\
TaxiBJ & TFT & 0.0913 / 0.1342 & 0.0956 / 0.1426 & 1 / 1 \\
M5 & XLinear & 0.1039 / 0.0877 & 0.1089 / 0.0950 & 1 / 1 \\
Rossmann & CEDAR & 0.1354 / 0.1375 & 0.1490 / 0.1600 & 1 / 1 \\
Seoul Bike & TFT & 0.2423 / 0.4053 & 0.2887 / 0.3732 & 1 / 3 \\
\bottomrule
\end{tabular}
\end{adjustbox}
\end{table}

\paragraph{Temporal consistency.}
We evaluate local changes using the mean absolute difference between predicted and observed adjacent increments,
\[
\operatorname{MAE}_{\Delta}=\operatorname{mean}_{n,h,f}
\frac{\left|(\widehat{Y}_{n,h,f}-\widehat{Y}_{n,h-1,f})-(Y_{n,h,f}-Y_{n,h-1,f})\right|}{\sigma_f},\qquad h=2,\ldots,H,
\]
where $n$ indexes test windows, $f$ indexes target channels, and $\sigma_f$ is the training-period scale used for prediction evaluation. Only adjacent steps within the same forecast window are compared. This metric measures recovery of local changes rather than smoothness alone. As shown in Figure~\ref{fig:appendix-temporal-changes}, Loki achieves lower mean change error than the best-overall baseline on four of the five datasets.

\begin{figure}[htbp]
\centering
\includegraphics[width=\linewidth]{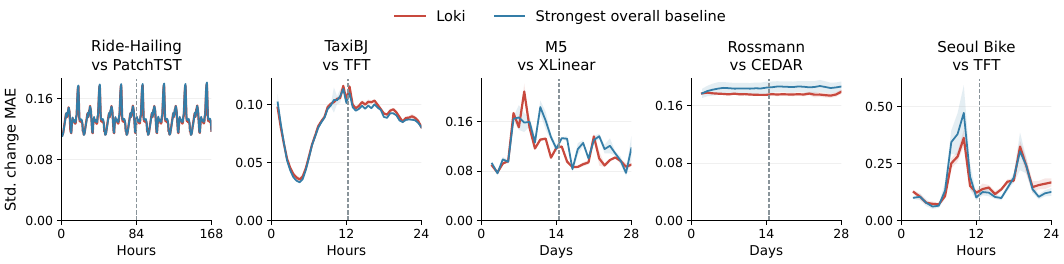}
\caption{\textcolor{black}{Error in changes between adjacent forecast steps. Lower standardized MAE between predicted and observed increments indicates more accurate recovery of local changes. Baselines are the same as in Figure~\ref{fig:appendix-horizon-steps}. Curves show the multi-seed mean, with shading indicating $\pm1$ standard deviation.}}
\label{fig:appendix-temporal-changes}
\end{figure}
\endgroup
\FloatBarrier

\Needspace{6\baselineskip}
\subsection{\textcolor{black}{Zero Targets and Zero Backbone Forecasts}}
\label{app:zero-targets}
\begingroup
\color{black}
All five datasets contain zero targets and no negative targets in the original target space. Table~\ref{tab:zero-target-prediction} reports the fraction of zero targets in the full data and prediction MAE at zero-target positions in the test set, before and after ERF restoration. Results are means and standard deviations across training seeds. The backbone forecast is the model output before future event responses are restored, not an observed event-free target.

\begin{table}[!htbp]
\centering
\color{black}
\caption{\textcolor{black}{Prediction at zero-target positions. Zero-target fractions use the full dataset; MAE is standardized and evaluated only at zero-target positions in the test set. Backbone and final predictions come from the same trained model.}}
\label{tab:zero-target-prediction}
\footnotesize
\setlength{\tabcolsep}{5pt}
\begin{adjustbox}{max width=\linewidth}
\begin{tabular}{lccc}
\toprule
Dataset & \makecell{Zero targets\\in full data} & \makecell{Backbone prediction MAE\\at zero targets} & \makecell{Final prediction MAE\\at zero targets} \\
\midrule
Ride-Hailing & 0.0106\% & $0.0638\pm0.0057$ & $0.0639\pm0.0056$ \\
TaxiBJ & 5.1286\% & $0.0049\pm0.0001$ & $0.0047\pm0.0001$ \\
M5 & 0.2626\% & \multicolumn{2}{c}{No zero targets in the test set} \\
Rossmann & 16.9946\% & $0.0219\pm0.0010$ & $0.0209\pm0.0010$ \\
Seoul Bike & 3.3676\% & $0.4310\pm0.1733$ & $0.4196\pm0.1650$ \\
\bottomrule
\end{tabular}
\end{adjustbox}
\end{table}

At zero-target positions, mean MAE decreases after response restoration on TaxiBJ, Rossmann, and Seoul Bike, while remaining approximately unchanged on Ride-Hailing. All zero targets in Seoul Bike correspond to non-operating hours (see Appendix~\ref{app:forecasting-protocol} for their distribution across splits); operating status is supplied as an environmental input describing whether the system provides service.

Exactly zero backbone forecasts occur only on Rossmann, with 0--5 cases per seed, all corresponding to zero targets. No zero-backbone/nonzero-target cases were observed. For a fixed backbone forecast, a finite positive response multiplier preserves zeros and signs: it cannot turn zero into a nonzero value, map a nonzero value exactly to zero, or reverse its sign. These constraints apply to the multiplicative response; zero historical observations do not force zero future forecasts, since the backbone can predict nonzero values from context.
\endgroup
\FloatBarrier

\Needspace{6\baselineskip}
\subsection{\textcolor{black}{Future New Events in Historical Adjustment}}
\label{app:future-events-history}
\begingroup
\color{black}
We test whether events starting at or after the forecast origin help historical adjustment. A relevant test window contains such a new event within the ERF support that can reach historical positions. Historical events continuing across the origin are not counted as new events. Table~\ref{tab:future-event-coverage} reports coverage across all five datasets. Coverage indicates an opportunity for this path to contribute, not evidence that the observed history contains an anticipatory effect.

\begin{table}[!htbp]
\centering
\small
\color{black}
\caption{\textcolor{black}{Test-window coverage of future new events that can contribute to the historical ERF response. The last column excludes weather events when identifying relevant windows.}}
\label{tab:future-event-coverage}
\begin{tabular}{lcc}
\toprule
Dataset & Relevant windows (\%) & \makecell{Relevant windows\\excluding weather (\%)} \\
\midrule
Ride-Hailing & 95.39 & 40.38 \\
TaxiBJ & 5.00 & 5.00 \\
M5 & 0.00 & 0.00 \\
Rossmann & 67.89 & 67.89 \\
Seoul Bike & 4.11 & 4.11 \\
\bottomrule
\end{tabular}
\end{table}

We select Ride-Hailing and Rossmann because they have the highest relevant-window coverage. We retain the complete schedule encoding and mask future new events' contributions only when computing the historical ERF response. Continuing historical events and the future ERF response remain unchanged. We retrain the ablated model using the same training settings and validation-based checkpoint selection as Full Loki, with matched training seeds. Shared parameters and parameter counts are unchanged.

\begin{table}[!htbp]
\centering
\small
\color{black}
\caption{\textcolor{black}{Removing future new events from historical adjustment and retraining. MAE is standardized and evaluated over the full forecast horizon of the selected windows. Values are multi-seed means; relative changes are computed within matched seed pairs and then averaged.}}
\label{tab:future-event-history-ablation}
\begin{adjustbox}{max width=\linewidth}
\begin{tabular}{llrrr}
\toprule
Dataset & Evaluation windows & Full Loki MAE & Ablated MAE & \makecell{Relative MAE\\change (\%)} \\
\midrule
\multirow{3}{*}{Ride-Hailing} & All windows & 0.156956 & 0.157156 & +0.128 \\
 & Event-relevant windows & 0.157565 & 0.157767 & +0.128 \\
 & Non-weather new-event windows & 0.170730 & 0.171032 & +0.177 \\
\midrule
\multirow{2}{*}{Rossmann} & All windows & 0.136492 & 0.136680 & +0.136 \\
 & Event-relevant windows & 0.135978 & 0.136138 & +0.117 \\
\bottomrule
\end{tabular}
\end{adjustbox}
\end{table}

The Ride-Hailing intervention includes weather events. Its non-weather new-event row is an evaluation subset of the same retrained model, not a separate intervention excluding weather. Table~\ref{tab:future-event-history-ablation} shows small average changes both overall and on event-relevant windows, indicating a small average forecasting benefit from this path within the evaluated scopes.
\endgroup
\FloatBarrier

\Needspace{6\baselineskip}
\subsection{\textcolor{black}{Sensitivity to Event-Schedule Errors}}
\label{app:schedule-errors}
\begingroup
\color{black}
We perturb event schedules in both historical and forecast windows while keeping trained models, targets, other covariates, and evaluation settings fixed. We test event deletion, separate start/end-time errors, and whole-event shifts at three levels. Whole-event shifts move both boundaries by the same number of steps, preserving duration and the intensity profile; they use the same timing levels and boundary constraints. Perturbations are consistent across entities and overlapping windows.

Tables~\ref{tab:schedule-timing-levels}--\ref{tab:schedule-affected-events} report nominal timing levels, relative prediction MAE changes, and actual affected-event counts. We average three perturbed schedules within each model seed, then report the mean and standard deviation across seeds.

Mean prediction error increases with deletion probability across the tested levels. Timing errors have asymmetric effects: Rossmann is more sensitive to start-time errors, whereas TaxiBJ is more sensitive to end-time errors. These experiments assess input-schedule errors, not changes in the underlying response mechanism.

\begin{table}[!htbp]
\centering
\footnotesize
\color{black}
\caption{\textcolor{black}{Nominal timing-error levels by event type. Median duration and displacement levels in steps use each dataset's sampling interval; the final column converts displacement to hours. Actual displacements can be smaller after boundary constraints.}}
\label{tab:schedule-timing-levels}
\setlength{\tabcolsep}{4pt}
\begin{adjustbox}{max width=\linewidth}
\begin{tabular}{llrrl}
\toprule
Dataset & Event type & \makecell{Median duration\\(steps)} & \makecell{Levels 1 / 2 / 3\\(steps)} & \makecell{Levels 1 / 2 / 3\\(hours)} \\
\midrule
\multirow{6}{*}{Ride-Hailing} & Public holidays & 144 & 36 / 72 / 144 & 18 / 36 / 72 \\
 & Social activity type 1 & 48 & 12 / 24 / 48 & 6 / 12 / 24 \\
 & Social activity type 2 & 48 & 12 / 24 / 48 & 6 / 12 / 24 \\
 & Traditional festivals & 48 & 12 / 24 / 48 & 6 / 12 / 24 \\
 & Weather factor & 5 & 2 / 3 / 5 & 1 / 1.5 / 2.5 \\
 & Western festivals & 48 & 12 / 24 / 48 & 6 / 12 / 24 \\
\midrule
\multirow{1}{*}{TaxiBJ} & Holiday type & 120 & 30 / 60 / 120 & 15 / 30 / 60 \\
\midrule
\multirow{7}{*}{M5} & Cultural events & 1 & 1 / 2 / 3 & 24 / 48 / 72 \\
 & National events & 1 & 1 / 2 / 3 & 24 / 48 / 72 \\
 & Religious events & 1 & 1 / 2 / 3 & 24 / 48 / 72 \\
 & Sporting events & 1 & 1 / 2 / 3 & 24 / 48 / 72 \\
 & Second same-day cultural event & 1 & 1 / 2 / 3 & 24 / 48 / 72 \\
 & Second same-day religious event & 1 & 1 / 2 / 3 & 24 / 48 / 72 \\
 & SNAP indicators & 2 & 1 / 2 / 3 & 24 / 48 / 72 \\
\midrule
\multirow{3}{*}{Rossmann} & Promotions & 5 & 2 / 3 / 5 & 48 / 72 / 120 \\
 & School holidays & 5 & 2 / 3 / 5 & 48 / 72 / 120 \\
 & Public holidays & 1 & 1 / 2 / 3 & 24 / 48 / 72 \\
\midrule
\multirow{1}{*}{Seoul Bike} & Holidays & 24 & 6 / 12 / 24 & 6 / 12 / 24 \\
\bottomrule
\end{tabular}
\end{adjustbox}
\par\smallskip
\begin{minipage}{\linewidth}
\scriptsize\color{black}
An event is a contiguous nonzero interval within a gap-free record; category changes split events, whereas dose changes do not. Timing levels are $\lceil0.25d\rceil$, $\lceil0.5d\rceil$, and $\lceil d\rceil$, where $d$ is the training median duration, adjusted to positive, strictly increasing integers. Directions are randomized. Displacements are clipped at neighboring-gap midpoints, data gaps, and record edges. Start/end errors skip single-sample events and retain at least one sample; expansion repeats endpoint doses and contraction crops the profile.
\end{minipage}
\end{table}

\begin{table}[!htbp]
\centering
\footnotesize
\color{black}
\caption{\textcolor{black}{Relative prediction MAE change (\%) against the same seed's original schedule; positive values indicate increased error. The same three perturbed schedules are averaged within each seed. Entries are means $\pm$ sample standard deviations across seeds, reflecting training variability rather than uncertainty across independent events. Deletion levels 1/2/3 correspond to 10\%/30\%/50\%; timing levels follow Table~\ref{tab:schedule-timing-levels}.}}
\label{tab:schedule-error-results}
\setlength{\tabcolsep}{4pt}
\begin{adjustbox}{max width=\linewidth}
\begin{tabular}{llccc}
\toprule
Dataset & Perturbation & Level 1 & Level 2 & Level 3 \\
\midrule
\multirow{4}{*}{Ride-Hailing} & Deletion & $+0.56\pm0.10$ & $+2.73\pm0.23$ & $+3.10\pm0.28$ \\
 & Start-time error & $+0.48\pm0.06$ & $+1.27\pm0.15$ & $+3.81\pm0.48$ \\
 & End-time error & $+0.97\pm0.08$ & $+2.66\pm0.26$ & $+6.19\pm0.66$ \\
 & Whole-event shift & $+1.06\pm0.06$ & $+1.98\pm0.13$ & $+3.00\pm0.19$ \\
\midrule
\multirow{4}{*}{TaxiBJ} & Deletion & $+0.00\pm0.00$ & $+1.00\pm0.55$ & $+2.00\pm1.10$ \\
 & Start-time error & $-0.13\pm0.30$ & $-0.93\pm0.68$ & $-1.27\pm1.61$ \\
 & End-time error & $+5.65\pm1.59$ & $+9.72\pm2.26$ & $+19.11\pm3.14$ \\
 & Whole-event shift & $+5.54\pm1.42$ & $+9.27\pm2.47$ & $+19.09\pm4.02$ \\
\midrule
\multirow{4}{*}{M5} & Deletion & $+5.48\pm0.55$ & $+10.86\pm1.46$ & $+12.71\pm1.45$ \\
 & Start-time error & $+8.82\pm0.95$ & $+13.43\pm1.63$ & $+17.32\pm2.18$ \\
 & End-time error & $+7.06\pm0.66$ & $+9.73\pm1.07$ & $+11.81\pm1.43$ \\
 & Whole-event shift & $+13.81\pm1.33$ & $+17.98\pm1.65$ & $+23.48\pm1.56$ \\
\midrule
\multirow{4}{*}{Rossmann} & Deletion & $+22.42\pm1.71$ & $+52.31\pm3.72$ & $+78.20\pm6.70$ \\
 & Start-time error & $+72.67\pm6.23$ & $+85.22\pm7.73$ & $+99.77\pm9.08$ \\
 & End-time error & $+16.70\pm2.54$ & $+28.82\pm3.67$ & $+47.20\pm4.76$ \\
 & Whole-event shift & $+104.92\pm11.15$ & $+124.78\pm14.33$ & $+152.13\pm19.23$ \\
\midrule
\multirow{4}{*}{Seoul Bike} & Deletion & $-0.30\pm0.06$ & $+0.36\pm0.40$ & $+0.93\pm0.88$ \\
 & Start-time error & $+0.20\pm0.09$ & $+0.52\pm0.16$ & $+1.36\pm0.29$ \\
 & End-time error & $-0.17\pm0.23$ & $-0.14\pm0.21$ & $+0.00\pm0.30$ \\
 & Whole-event shift & $+0.02\pm0.33$ & $+0.36\pm0.41$ & $+1.03\pm0.65$ \\
\bottomrule
\end{tabular}
\end{adjustbox}
\end{table}

\begin{table}[!htbp]
\centering
\footnotesize
\color{black}
\caption{\textcolor{black}{Actual affected events intersecting the test windows' historical or forecast contexts. Each cell gives the minimum--maximum count across three perturbed schedules; a single number indicates identical counts. Counts are deduplicated modified label intervals, not necessarily distinct real-world incidents, and are not multiplied by model seeds. Levels follow Table~\ref{tab:schedule-error-results}.}}
\label{tab:schedule-affected-events}
\setlength{\tabcolsep}{5pt}
\begin{adjustbox}{max width=\linewidth}
\begin{tabular}{llrrr}
\toprule
Dataset & Perturbation & Level 1 & Level 2 & Level 3 \\
\midrule
\multirow{4}{*}{Ride-Hailing} & Deletion & 4,032--4,209 & 12,347--12,511 & 20,567--20,815 \\
 & Start-time error & 37,181--37,183 & 37,181--37,183 & 37,181--37,183 \\
 & End-time error & 33,266--33,401 & 33,266--33,401 & 33,266--33,401 \\
 & Whole-event shift & 37,010--37,157 & 37,010--37,157 & 37,010--37,157 \\
\midrule
\multirow{4}{*}{TaxiBJ} & Deletion & 0 & 0--1 & 0--1 \\
 & Start-time error & 1 & 1 & 1 \\
 & End-time error & 1 & 1 & 1 \\
 & Whole-event shift & 1 & 1 & 1 \\
\midrule
\multirow{4}{*}{M5} & Deletion & 18--22 & 49--63 & 85--102 \\
 & Start-time error & 105 & 105 & 105 \\
 & End-time error & 63--70 & 63--70 & 63--70 \\
 & Whole-event shift & 127--139 & 127--139 & 127--139 \\
\midrule
\multirow{4}{*}{Rossmann} & Deletion & 5--16 & 33--40 & 62--70 \\
 & Start-time error & 119 & 119 & 119 \\
 & End-time error & 115--118 & 115--118 & 115--118 \\
 & Whole-event shift & 140--144 & 140--144 & 140--144 \\
\midrule
\multirow{4}{*}{Seoul Bike} & Deletion & 0--1 & 0--3 & 0--3 \\
 & Start-time error & 3 & 3 & 3 \\
 & End-time error & 3 & 3 & 3 \\
 & Whole-event shift & 3 & 3 & 3 \\
\bottomrule
\end{tabular}
\end{adjustbox}
\end{table}

\endgroup
\FloatBarrier

\subsection{\textcolor{black}{Diagnostics of Post-Event Extrapolation}}
\label{app:post-event-extrapolation}
\begingroup
\color{black}

\paragraph{Diagnostic.}
Event end does not imply effect end. We select windows where the historical event has ended and the observed deviation from a same-entity, same-time-of-week non-event reference has fallen to at most half its historical level. We then identify forecasts that extend the historical anomaly beyond the observed residual. This criterion allows genuine residual effects rather than assuming that the response immediately returns to zero. Table~\ref{tab:extrapolation-rates} reports the complete results for 19 models. M5 has no eligible window, and Seoul Bike has only one eligible sample; rate comparisons therefore cover Ride-Hailing, TaxiBJ and Rossmann.

\begin{table}[!htbp]
\centering
\small
\color{black}
\caption{\textcolor{black}{Erroneous-extrapolation diagnostic rates. Each column uses the same eligible window--target pairs for all 19 models: 1,289 on Ride-Hailing, 1,182 on TaxiBJ and 10,695 on Rossmann.}}
\label{tab:extrapolation-rates}
\begin{tabular}{lrrr}
\toprule
Model & Ride-Hailing & TaxiBJ & Rossmann \\
\midrule
DLinear & 33.4\% & 95.5\% & 19.9\% \\
PatchTST & 18.2\% & 93.1\% & 7.2\% \\
SparseTSF & 38.9\% & 94.6\% & 14.0\% \\
TimeMixer & 21.7\% & 98.0\% & 10.7\% \\
iTransformer & 20.0\% & 94.7\% & 6.3\% \\
TimeXer & 22.6\% & 97.1\% & 9.1\% \\
CrossLinear & 21.2\% & 97.5\% & 8.3\% \\
XLinear & 25.4\% & 97.9\% & 8.6\% \\
TiDE & 17.6\% & 97.6\% & 11.8\% \\
DAG & 20.2\% & 97.7\% & 5.3\% \\
TFT & 7.7\% & 71.0\% & 8.1\% \\
AutoARIMA-X & 36.7\% & 0.1\% & 11.6\% \\
Log-linear & 19.3\% & 74.4\% & 13.2\% \\
Prophet & 37.5\% & 1.5\% & 23.5\% \\
NeuralProphet & 26.5\% & 95.1\% & 14.8\% \\
NBEATSx & 17.0\% & 98.4\% & 14.6\% \\
SPADE & 15.1\% & 47.8\% & 9.7\% \\
CEDAR & 19.7\% & 56.6\% & 5.1\% \\
\textbf{Loki} & \textbf{14.4\%} & \textbf{13.9\%} & \textbf{8.0\%} \\
\bottomrule
\end{tabular}
\par\smallskip
\begin{minipage}{\linewidth}\footnotesize
The reference is the median of at least two non-event, operating observations at the same weekly slot within four weeks before or after the evaluated time. Future observations are used only for retrospective diagnosis, never as model inputs. Eligible pairs have reference coverage of at least 50\% in the historical event interval and post-event positions, with a historical deviation exceeding both 0.05 training standard deviations and 10\% of the reference level. In the historical-deviation direction, the mean forecast deviations from both the actual observations and the reference must exceed the larger of 0.05 training standard deviations and 10\% of the historical deviation. Stochastic models require agreement in at least four of five seeds; deterministic models use one fitted result. TaxiBJ's eligible windows all come from 2016-04-05, after Qingming. These output patterns alone do not establish that historical inputs caused the error.
\end{minipage}
\end{table}

\paragraph{Controlled history replacement.}
We freeze the models and all future inputs, replacing only historical event-period observations with a normal reference constructed from pre-origin observations. Two replacements at non-event positions serve as controls. A window--target pair qualifies when same-direction bias, MAE and MSE all decrease in at least four of five seeds, and the MAE improvement exceeds both controls. Table~\ref{tab:history-replacement-counts} reports the qualifying counts. These comparisons support historical event inputs as a contributor to erroneous extrapolation in the qualifying samples; the replacement reference is not causal ground truth.

\begin{table}[!htbp]
\centering
\small
\color{black}
\caption{\textcolor{black}{Window--target pairs supporting historical-input contribution under controlled replacement. The selected pairs form a smaller, data-selected subset of the diagnostic sample in Table~\ref{tab:extrapolation-rates}.}}
\label{tab:history-replacement-counts}
\begin{tabular}{lrrrrr}
\toprule
Dataset & Selected pairs & DAG & TiDE & TFT & Loki \\
\midrule
Ride-Hailing & 67 & 6 & 1 & 0 & 2 \\
TaxiBJ & 63 & 18 & 15 & 0 & 0 \\
Rossmann & 32 & 0 & 2 & 0 & 0 \\
\bottomrule
\end{tabular}
\par\smallskip
\begin{minipage}{\linewidth}\footnotesize
We select up to 32 windows per dataset, stratified by forecast date without using model errors. Replacement affects the most recent historical event interval; event indicators remain unchanged. One control replaces the same number of non-event observations using the pre-origin reference; the other transfers the event-replacement changes to non-event positions, preserving their count and per-target signed total. Both preferentially match time of day. A qualifying seed must satisfy the original output diagnostic, reduce same-direction mean error by more than 0.01 training standard deviations and MAE by more than 0.001, reduce MSE, and improve MAE over each control by more than 0.005. The criterion must hold in at least four of five seeds.
\end{minipage}
\end{table}

\paragraph{Post-event prediction accuracy.}
A low erroneous-extrapolation rate alone does not establish accurate forecasts. \textcolor{black}{\hypersetup{linkcolor=black}Table~\ref{tab:post-event-prediction} therefore reports prediction MAE during the first 24 hours after events across all five datasets.} Loki ranks first on Ride-Hailing and Rossmann, second on M5, third on TaxiBJ and fifth on Seoul Bike.

\begin{table}[!htbp]
\centering
\footnotesize
\color{black}
\caption{\textcolor{black}{Prediction MAE within 24 hours after events. Stochastic models report mean MAE $\pm$ sample standard deviation across training seeds; deterministic models report one fitted result. The best and second-best means in each dataset are bold and underlined, respectively, based on unrounded values.}}
\label{tab:post-event-prediction}
\setlength{\tabcolsep}{3pt}
\renewcommand{\arraystretch}{1.22}
\begin{adjustbox}{max width=\linewidth}
\begin{tabular}{lccccc}
\toprule
Model & Ride-Hailing & TaxiBJ & M5 & Rossmann & Seoul Bike \\
\midrule
DLinear & $0.1706 \pm 0.0019$ & $0.1703 \pm 0.0029$ & $0.0998 \pm 0.0050$ & $0.2207 \pm 0.0015$ & $0.9176 \pm 0.0594$ \\
PatchTST & $\underline{0.1397} \pm 0.0006$ & $0.1772 \pm 0.0155$ & $0.0990 \pm 0.0056$ & $0.1817 \pm 0.0023$ & $0.8783 \pm 0.0518$ \\
SparseTSF & $0.1643 \pm 0.0003$ & $0.1664 \pm 0.0064$ & $\mathbf{0.0864} \pm 0.0030$ & $0.1983 \pm 0.0015$ & $0.8787 \pm 0.0357$ \\
TimeMixer & $0.1434 \pm 0.0004$ & $0.1896 \pm 0.0155$ & $0.1025 \pm 0.0082$ & $0.1824 \pm 0.0030$ & $0.8502 \pm 0.0459$ \\
iTransformer & $0.1454 \pm 0.0006$ & $0.1723 \pm 0.0193$ & $0.0977 \pm 0.0047$ & $0.1875 \pm 0.0021$ & $0.9827 \pm 0.0290$ \\
TimeXer & $0.1489 \pm 0.0009$ & $0.1924 \pm 0.0294$ & $0.1018 \pm 0.0043$ & $0.1878 \pm 0.0012$ & $0.8923 \pm 0.0307$ \\
CrossLinear & $0.1427 \pm 0.0003$ & $0.1766 \pm 0.0097$ & $0.0939 \pm 0.0067$ & $0.1842 \pm 0.0029$ & $0.9355 \pm 0.0253$ \\
XLinear & $0.1438 \pm 0.0005$ & $0.1863 \pm 0.0249$ & $0.0951 \pm 0.0046$ & $0.1845 \pm 0.0033$ & $0.9410 \pm 0.0122$ \\
TiDE & $0.1453 \pm 0.0023$ & $0.1992 \pm 0.0170$ & $0.0921 \pm 0.0084$ & $0.1499 \pm 0.0032$ & $0.8344 \pm 0.1218$ \\
DAG & $0.1439 \pm 0.0008$ & $0.1888 \pm 0.0122$ & $0.1001 \pm 0.0064$ & $\underline{0.1374} \pm 0.0032$ & $0.8962 \pm 0.0981$ \\
TFT & $0.1520 \pm 0.0016$ & $0.1237 \pm 0.0112$ & $0.0984 \pm 0.0054$ & $0.1578 \pm 0.0079$ & $0.3850 \pm 0.0187$ \\
AutoARIMA-X & $0.3447$ & $0.1891$ & $0.1192$ & $0.1972$ & $0.7312$ \\
Log-linear & $0.1908$ & $\mathbf{0.1059}$ & $0.1006$ & $0.2146$ & $\mathbf{0.3615}$ \\
Prophet & $0.2363$ & $0.2088$ & $0.0885$ & $0.2114$ & $1.1959$ \\
NeuralProphet & $0.1696 \pm 0.0052$ & $0.1976 \pm 0.0255$ & $0.0918 \pm 0.0055$ & $0.2124 \pm 0.0153$ & $1.7984 \pm 0.5591$ \\
NBEATSx & $0.1526 \pm 0.0005$ & $0.1852 \pm 0.0090$ & $0.1054 \pm 0.0083$ & $0.1868 \pm 0.0020$ & $0.8357 \pm 0.1054$ \\
SPADE & $0.1666 \pm 0.0010$ & $0.2680 \pm 0.0097$ & $0.1002 \pm 0.0058$ & $0.2091 \pm 0.0114$ & $0.4989 \pm 0.0610$ \\
CEDAR & $0.2168 \pm 0.0275$ & $\underline{0.1148} \pm 0.0188$ & $0.1001 \pm 0.0033$ & $0.1662 \pm 0.0043$ & $\underline{0.3836} \pm 0.0394$ \\
\textbf{Loki} & $\mathbf{0.1392} \pm 0.0002$ & $0.1181 \pm 0.0097$ & $\underline{0.0880} \pm 0.0036$ & $\mathbf{0.1318} \pm 0.0021$ & $0.5013 \pm 0.0487$ \\
\bottomrule
\end{tabular}
\end{adjustbox}
\par\smallskip
\begin{minipage}{\linewidth}\footnotesize
We evaluate pure post-event positions, excluding active-event positions and positions also in another event's pre-event interval. This schedule-based grouping does not apply the demand-recovery filter in Table~\ref{tab:extrapolation-rates}. For daily data, 24 hours corresponds to one step. Evaluation weights forecast windows, so the same real timestamp may recur. M5's post-event positions cover only one date.
\end{minipage}
\end{table}

\endgroup
\FloatBarrier

\subsection{\textcolor{black}{Initialization and Temporal Bases}}
\label{app:initialization-temporal-bases}
\begingroup
\color{black}
We compare response initialization and temporal-basis settings. Random initialization draws the response weights from $\mathcal{N}(0,0.01^2)$. The simplified basis bank retains only the impulse and the slowest forward/backward pair, with the response support unchanged.

\begin{table}[!htbp]
\centering
\small
\color{black}
\caption{\textcolor{black}{Prediction MAE under response-initialization and temporal-basis variants. Values are multi-seed means $\pm$ sample standard deviations.}}
\label{tab:initialization-temporal-bases}
\begin{tabular}{lcc}
\toprule
Variant & Ride-Hailing prediction MAE & M5 prediction MAE \\
\midrule
Loki & $0.15696\pm0.00031$ & $0.09583\pm0.00109$ \\
Random response initialization & $0.15689\pm0.00035$ & $0.09621\pm0.00172$ \\
Simplified temporal bases & $0.15782\pm0.00012$ & $0.09823\pm0.00230$ \\
\bottomrule
\end{tabular}
\end{table}

Random and zero initialization give similar prediction accuracy. We retain zero initialization so that Lift Out and Kick In start as identity transformations. Reducing the basis count from 19 to 3 on Ride-Hailing and from 7 to 3 on M5 increases MAE on both datasets, supporting the multi-scale bidirectional basis bank.
\endgroup
\FloatBarrier

\subsection{\textcolor{black}{Statistical Significance}}
\label{app:statistical-significance}
\begingroup
\color{black}
For each dataset, we compare Loki with the non-Loki model having the lowest mean prediction MAE. We average each model's standardized absolute errors across training seeds, then aggregate their difference by test forecast origin. We apply a two-sided Diebold--Mariano (DM) test \textcolor{black}{\hypersetup{citecolor=black}\citep{Diebold01071995}} with Newey--West HAC variance estimation \textcolor{black}{\hypersetup{citecolor=black}\citep{newey1986simple}} to account for overlapping forecast windows. HAC lags are 7, 1, 28, and 1 origins for Ride-Hailing, TaxiBJ, Rossmann, and Seoul Bike, respectively. The test uses a Student-$t$ reference with $n-1$ degrees of freedom, where $n$ is the number of origins. Seed wins compare Loki with the baseline at the same seed.

\begin{table}[!htbp]
\centering
\footnotesize
\color{black}
\caption{\textcolor{black}{Loki versus the strongest baseline on each dataset. Prediction MAE follows baseline $\rightarrow$ Loki; reductions are relative to the baseline. Origin wins use seed-averaged losses, while seed wins use matched training seeds.}}
\label{tab:statistical-significance}
\setlength{\tabcolsep}{3pt}
\begin{adjustbox}{max width=\linewidth}
\begin{tabular}{llcrrrcc}
\toprule
Dataset & Baseline & Prediction MAE & \shortstack{MAE reduction\\(\%)} & \shortstack{Test\\origins} & \shortstack{Origin wins\\(\%)} & DM $p$ & \shortstack{Seed\\wins} \\
\midrule
Ride-Hailing & PatchTST & $0.1618\rightarrow0.1570$ & 3.0 & 178 & 80 & $0.010$ & 5/5 \\
TaxiBJ & TFT & $0.1191\rightarrow0.1128$ & 5.3 & 20 & 80 & $0.20$ & 4/5 \\
M5 & XLinear & $0.1019\rightarrow0.0958$ & 6.0 & 1 & --- & N/A & 5/5 \\
Rossmann & CEDAR & $0.1545\rightarrow0.1365$ & 11.7 & 162 & 98 & $<0.001$ & 5/5 \\
Seoul Bike & TFT & $0.3310\rightarrow0.3238$ & 2.2 & 73 & 59 & $0.85$ & 3/5 \\
\bottomrule
\end{tabular}
\end{adjustbox}
\end{table}

Loki's advantage is statistically significant on Ride-Hailing and Rossmann at the 0.05 level, but not on TaxiBJ or Seoul Bike. M5 has only one test forecast origin, so this origin-based DM test is not applicable.
\endgroup
\FloatBarrier

\section{Controlled Event-Effect Recovery Details}
\label{app:controlled-recovery}

\subsection{Semi-Synthetic Event Construction}
\label{app:semi-synthetic-construction}

\textcolor{black}{We construct semi-synthetic versions of Ride-Hailing and M5 using the original series as a reference. We test multiplicative, additive, and state-dependent responses, together with a null control. For the multiplicative setting,}
\[
Y_{i,t,f}
=
Y^{\mathrm{ref}}_{i,t,f}
\exp\!\left(S^*_{i,t,f}\right),
\]
where $i$, $t$, and $f$ index forecasting unit, time, and target variable. The reference series retains its original temporal dynamics, noise, covariates, and native event-related variation; the experiment adds only the controlled artificial response $S^*$.

Each forecasting unit receives non-overlapping artificial events $j\in\mathcal E_i$. Event $j$ has type $k_j\in\{+,-\}$, corresponding to uplift and suppression \textcolor{black}{in the multiplicative setting}, onset $o_j$, duration $\ell_j$, and dose $q_j\in\{0.5,1.0,1.5\}$. Its injected log response is
\begin{equation}
\label{eq:app-injected-response}
S^*_{i,t,f}
=
\sum_{j\in\mathcal E_i}
a_{i,k_j}\,b_{k_j,f}\,A_{k_j}\,q_j\,
h_{k_j}(t;o_j,\ell_j),
\end{equation}
where $a_{i,k}\sim U(0.7,1.3)$ is a forecasting-unit- and event-type-specific gain. For Ride-Hailing, $b_{k,f}\sim U(0.85,1.15)$ is an event-type- and target-specific gain; M5 has one target and uses $b_{k,f}=1$. The base log amplitudes are $A_+=\log(1.3)$ and $A_-=\log(0.75)$.

Let $\ell_{\mathrm{pre}}$ and $\ell_{\mathrm{post}}$ denote the anticipation and post-event support lengths. Uplift uses a linear anticipatory ramp, a constant response during the event interval, and a linear decay after the event:
\begin{equation}
\label{eq:app-uplift-profile}
h_+(t;o,\ell)
=
\begin{cases}
\dfrac{t-o+\ell_{\mathrm{pre}}+1}{\ell_{\mathrm{pre}}+1},
& o-\ell_{\mathrm{pre}}\le t<o,\\[4pt]
1, & o\le t<o+\ell,\\[4pt]
\dfrac{o+\ell+\ell_{\mathrm{post}}-t}{\ell_{\mathrm{post}}+1},
& o+\ell\le t<o+\ell+\ell_{\mathrm{post}},\\[4pt]
0, & \text{otherwise}.
\end{cases}
\end{equation}
Suppression has no anticipatory component; its response decreases linearly from $1$ to $0.5$ during the event and then enters a \textcolor{black}{lower-amplitude} tail beginning at one quarter of the base response:
\begin{equation}
\label{eq:app-suppression-profile}
h_-(t;o,\ell)
=
\begin{cases}
1-\dfrac{t-o}{2(\ell-1)}, & o\le t<o+\ell,\\[4pt]
\dfrac{1}{4}
\dfrac{o+\ell+\ell_{\mathrm{post}}-t}{\ell_{\mathrm{post}}},
& o+\ell\le t<o+\ell+\ell_{\mathrm{post}},\\[4pt]
0, & \text{otherwise}.
\end{cases}
\end{equation}
Table~\ref{tab:semi-synthetic-settings} summarizes the dataset-specific temporal settings.

\begin{table}[t]
\centering
\caption{Semi-synthetic event-generation settings used for controlled recovery.}
\label{tab:semi-synthetic-settings}
\small
\begin{adjustbox}{max width=\columnwidth}
\begin{tabular}{lcc}
\toprule
Setting & Ride-Hailing & M5 \\
\midrule
Temporal resolution & 30 minutes & 1 day \\
Event duration $\ell$ & 12 / 24 / 48 steps & 3 / 5 / 7 days \\
Uplift anticipation $\ell_{\mathrm{pre}}$ & 4 steps & 1 day \\
Post-event tail $\ell_{\mathrm{post}}$ & 12 steps & 2 days \\
Gap between response supports & 48--144 steps & 7--14 days \\
\bottomrule
\end{tabular}
\end{adjustbox}
\end{table}

For each forecasting unit, events are scheduled independently within the original train, validation, and test splits. Artificial response supports neither overlap nor cross split boundaries. Event type, duration, and dose are sampled uniformly from their respective choices, and the gap between consecutive supports is sampled uniformly over the integer range in Table~\ref{tab:semi-synthetic-settings}. Missing dates are handled by generating on the complete temporal grid and then mapping back to observed rows; no values are imputed and no additional observation noise is introduced.

\textcolor{black}{We generate three independent semi-synthetic configurations. Within each configuration, schedules, doses, and gains are fixed across training seeds and compared models. The model receives two artificial event channels. For an event of type $k$, the corresponding input is $e_{i,t,k}=q_j$ during $o_j\le t<o_j+\ell_j$ and zero otherwise. Anticipatory and post-event effects are not directly encoded in these inputs. The response profiles, gains, and injected effects are never provided as model inputs or training supervision.}

\begingroup
\color{black}
\paragraph{Additive injection.}
Let $S^*_{i,t,k,f}$ denote the contribution of event type $k$ to the multiplicative log response defined above, and let $\mu^{\mathrm{ref}}_{i,f}$ be the mean reference target for unit $i$ and target $f$ in the training split. Using the same schedules and temporal profiles, we inject
\begin{equation}
\label{eq:app-additive-injection}
\Delta^*_{i,t,f}=\mu^{\mathrm{ref}}_{i,f}\sum_k\left(\exp\!\left|S^*_{i,t,k,f}\right|-1\right),
\qquad Y_{i,t,f}=Y^{\mathrm{ref}}_{i,t,f}+\Delta^*_{i,t,f}.
\end{equation}
The increment depends on the training reference mean rather than the instantaneous reference level. Both channels produce nonnegative increments in this setting, preserving nonnegative targets; this experiment therefore tests additive uplift rather than signed additive effects. State-dependent multiplicative injection is specified in Appendix~\ref{app:state-dependent-recovery}.
\endgroup

\subsection{\textcolor{black}{Experimental Conditions}}
\label{app:recovery-conditions}
\begingroup
\color{black}
Full Loki uses a shared multiplicative ERF. Independent ERFs use separate response functions for historical adjustment and future reconstruction. Half-independent reduces their rank and number of temporal bases to approximately match the shared ERF's total parameter budget. Additive Loki retains the shared ERF and backbone, but subtracts the estimated increment from history and adds it to the backbone prediction. In the semi-synthetic additive model, the raw ERF output is multiplied by the reference training scale $\sigma_f^{\mathrm{ref}}$ to express its output in target units. All models receive the same artificial event descriptors and native covariates. Models are trained separately for each injection setting using only the forecasting loss.

The null control retains the same artificial schedules and doses but leaves the target equal to $Y^{\mathrm{ref}}$, making the true artificial increment zero. The no-recovery references leave the historical series unadjusted or set the estimated future event increment to zero. These references are computed directly from the injected truth and do not require training a model with event descriptors removed.

Each configuration uses three training seeds. We first average over training seeds within a configuration, then report the mean and sample standard deviation across the three generated configurations. Checkpoints are selected by validation forecasting MAE. Training normalization is fitted on the training portion of each injected dataset; all reported recovery and prediction errors are evaluated using the common pre-injection training scale $\sigma_f^{\mathrm{ref}}$.
\endgroup

\subsection{\textcolor{black}{Recovery and Prediction Metrics}}
\label{app:recovery-metrics}
\begingroup
\color{black}
Let $\Delta^*=Y-Y^{\mathrm{ref}}$ be the known artificial target increment. For a multiplicative model, let $\widehat S_{\mathrm{syn}}$ be the sum of the estimated responses for the two artificial event channels. For an additive model, let $\widehat A_{\mathrm{syn}}$ be their estimated increment in target units. We evaluate
\begin{equation}
\label{eq:app-history-restoration}
\widetilde Y^{\mathrm{hist}}=
\begin{cases}
Y^{\mathrm{hist}}\odot\exp(-\widehat S^{\mathrm{hist}}_{\mathrm{syn}}), & \text{multiplicative model},\\
Y^{\mathrm{hist}}-\widehat A^{\mathrm{hist}}_{\mathrm{syn}}, & \text{additive model},
\end{cases}
\end{equation}
and
\begin{equation}
\label{eq:app-effective-artificial-response}
\widehat\Delta^{\mathrm{fut}}=
\begin{cases}
Y^{\mathrm{ref,fut}}\odot\left[\exp(\widehat S^{\mathrm{fut}}_{\mathrm{syn}})-1\right], & \text{multiplicative model},\\
\widehat A^{\mathrm{fut}}_{\mathrm{syn}}, & \text{additive model}.
\end{cases}
\end{equation}
These calculations isolate the artificial response and retain native event-associated variation in the reference series. The reference future level is used only to evaluate response recovery, not to produce the forecast. Both model forms are therefore evaluated in the same target-increment space without conflating effect recovery with errors in the backbone's predicted level.

For each target, MAE averages the absolute error divided by $\sigma_f^{\mathrm{ref}}$ over eligible window--time positions; the final metric averages equally across targets with eligible positions. Historical and future recovery use the respective portions of the artificial response support, including anticipation and post-event tails. Positions affected by an event whose onset or end is not strictly inside the joint history--forecast window are excluded. Reference targets equal to zero are retained. The null control uses the support of the same event schedules even though no effect is injected.

\begin{itemize}
\item \textbf{Historical MAE before adjustment}: error between $Y^{\mathrm{hist}}$ and $Y^{\mathrm{ref,hist}}$, measuring the injected historical variation.
\item \textbf{Historical MAE after adjustment}: error between $\widetilde Y^{\mathrm{hist}}$ and $Y^{\mathrm{ref,hist}}$.
\item \textbf{Future effect MAE}: error between $\widehat\Delta^{\mathrm{fut}}$ and $\Delta^{*\mathrm{fut}}$.
\item \textbf{Zero-response future effect MAE}: error when $\widehat\Delta^{\mathrm{fut}}=0$.
\item \textbf{Response-support prediction MAE}: error between the final forecast and the injected target on the same future positions used for effect recovery.
\item \textbf{Overall prediction MAE}: error between the final forecast and the injected target over all test forecast positions.
\end{itemize}
\endgroup

\subsection{\textcolor{black}{Recovery Results}}
\label{app:recovery-results}
\begingroup
\color{black}
Table~\ref{tab:recovery-reference} reports the no-recovery references, and Table~\ref{tab:controlled-event-recovery} extends the main-paper recovery table with prediction errors and uncertainty across generated configurations.

\begin{table}[htbp]
\centering
\small
\color{black}
\caption{\textcolor{black}{No-recovery references for the injected effects. Values are standardized MAE, reported as means $\pm$ sample standard deviations across generated configurations. Both references are zero in the null control.}}
\label{tab:recovery-reference}
\begin{tabular}{llcc}
\toprule
Dataset & Injection & Historical MAE & Zero-response future \\
 & & before adjustment & effect MAE \\
\midrule
\multirow{2}{*}{Ride-Hailing} & Multiplicative & $0.3333\pm0.0054$ & $0.3284\pm0.0046$ \\
 & Additive & $0.3176\pm0.0050$ & $0.3138\pm0.0045$ \\
\midrule
\multirow{2}{*}{M5} & Multiplicative & $0.1927\pm0.0083$ & $0.2493\pm0.0288$ \\
 & Additive & $0.1825\pm0.0091$ & $0.2003\pm0.0230$ \\
\bottomrule
\end{tabular}
\end{table}

\input{table/event}

Under multiplicative injection, shared Loki has lower historical and future effect MAE than Independent ERFs, Half-independent, and Additive Loki on both datasets. Under additive injection, multiplicative Loki reduces future effect MAE from 0.3138 to 0.1004 on Ride-Hailing and from 0.2003 to 0.0552 on M5; the matched additive model achieves 0.0247 and 0.0141. The corresponding prediction errors also favor the additive model. Thus, the multiplicative structure recovers part of the additive effects but does not match the recovery accuracy of the additive structure.

For an additional prediction reference under additive injection, we retain the trained multiplicative model's historical adjustment, backbone prediction, and native event response, and disable only the future artificial response without retraining. Its response-support prediction MAE is $0.3778\pm0.0039$ on Ride-Hailing and $0.2075\pm0.0202$ on M5, compared with $0.1869\pm0.0011$ and $0.1191\pm0.0032$ for the complete multiplicative model. This prediction reference differs from the zero-response effect reference: it retains the model's predicted underlying trajectory.

In the null control, historical and future effect MAE remain below 0.01 on both datasets. The nonzero errors indicate limited spurious attribution. Recovery of injected effects under these controlled conditions does not establish identifiability or causal interpretation for real event effects.
\endgroup

\subsection{\textcolor{black}{State-Dependent Responses and Event Interactions}}
\label{app:state-dependent-recovery}
\label{app:overlap-interaction-recovery}
\begingroup
\color{black}
\paragraph{State-dependent responses.}
We keep the multiplicative setting's schedules and temporal profiles, but vary each event's response factor with the recent reference-series level. For each unit, we first average the reference targets after dividing each by $\sigma_f^{\mathrm{ref}}$. At the start of an event's response support, we compute the mean of this signal over the preceding week (336 half-hour steps for Ride-Hailing and 7 daily steps for M5), excluding the current position. We standardize this recent mean using the mean and standard deviation of the corresponding rolling means in the unit's training split, then apply $\tanh$ to obtain $z_{i,j}\in[-1,1]$. The event's injected log response is multiplied by $1+0.5z_{i,j}$, with this factor held fixed throughout its response support:
\begin{equation}
\label{eq:app-state-injection}
S^{*\mathrm{state}}_{i,t,f}
=\sum_j(1+0.5z_{i,j})\,a_{i,k_j}b_{k_j,f}A_{k_j}q_j h_{k_j}(t;o_j,\ell_j),
\qquad
Y_{i,t,f}=Y^{\mathrm{ref}}_{i,t,f}\exp(S^{*\mathrm{state}}_{i,t,f}).
\end{equation}
The state variable is used only by the data generator and is not added to the ERF inputs.

\begin{table}[!htbp]
\centering
\footnotesize
\color{black}
\caption{\textcolor{black}{Prediction and recovery MAE under state-dependent responses and overlapping events, using shared multiplicative Loki. Values are means $\pm$ standard deviations across generated configurations; aggregation and metrics follow Appendices~\ref{app:recovery-conditions}--\ref{app:recovery-metrics}. ``RMS-matched'' denotes the no-interaction condition matched to the interaction condition's training perturbation magnitude.}}
\label{tab:state-dependent-recovery}
\label{tab:overlap-interaction-recovery}
\setlength{\tabcolsep}{3pt}
\renewcommand{\arraystretch}{1.35}
\begin{adjustbox}{max width=\textwidth}
\begin{tabular}{llcccccc}
\toprule
\multirow{2}{*}{Dataset} & \multirow{2}{*}{Condition} & \multicolumn{2}{c}{Prediction MAE} & \multicolumn{2}{c}{Historical MAE} & \multicolumn{2}{c}{Future effect MAE} \\
\cmidrule(lr){3-4}\cmidrule(lr){5-6}\cmidrule(lr){7-8}
 & & Overall & \shortstack{Response\\support} & \shortstack{Before\\adjustment} & \shortstack{After\\adjustment} & \shortstack{Zero\\response} & \shortstack{Recovered\\response} \\
\midrule
\multirow{6}{*}{Ride-Hailing} & State-dependent & $0.1886 \pm 0.0009$ & $0.2464 \pm 0.0030$ & $0.5040 \pm 0.0105$ & $0.1339 \pm 0.0027$ & $0.4965 \pm 0.0091$ & $0.1510 \pm 0.0040$ \\
\cmidrule(lr){2-8}
 & Count-matched, separate & $0.1616 \pm 0.0003$ & $0.1694 \pm 0.0015$ & $0.3333 \pm 0.0054$ & $0.0207 \pm 0.0004$ & $0.3284 \pm 0.0046$ & $0.0228 \pm 0.0007$ \\
 & Count-matched, overlap & $0.1609 \pm 0.0001$ & $0.1704 \pm 0.0002$ & $0.2854 \pm 0.0038$ & $0.0240 \pm 0.0004$ & $0.2827 \pm 0.0041$ & $0.0261 \pm 0.0006$ \\
\cmidrule(lr){2-8}
 & Overlap, no interaction & $0.1625 \pm 0.0004$ & $0.1703 \pm 0.0016$ & $0.3024 \pm 0.0046$ & $0.0227 \pm 0.0006$ & $0.2986 \pm 0.0037$ & $0.0247 \pm 0.0007$ \\
 & Overlap, with interaction & $0.1661 \pm 0.0004$ & $0.1799 \pm 0.0018$ & $0.3185 \pm 0.0065$ & $0.0421 \pm 0.0012$ & $0.3142 \pm 0.0055$ & $0.0438 \pm 0.0012$ \\
 & Overlap, RMS-matched & $0.1627 \pm 0.0005$ & $0.1708 \pm 0.0020$ & $0.3110 \pm 0.0054$ & $0.0232 \pm 0.0008$ & $0.3071 \pm 0.0045$ & $0.0253 \pm 0.0013$ \\
\midrule
\multirow{6}{*}{M5} & State-dependent & $0.1088 \pm 0.0051$ & $0.1392 \pm 0.0103$ & $0.2383 \pm 0.0090$ & $0.0445 \pm 0.0032$ & $0.3311 \pm 0.0363$ & $0.0704 \pm 0.0088$ \\
\cmidrule(lr){2-8}
 & Count-matched, separate & $0.1022 \pm 0.0009$ & $0.1117 \pm 0.0014$ & $0.1713 \pm 0.0029$ & $0.0157 \pm 0.0008$ & $0.1929 \pm 0.0040$ & $0.0177 \pm 0.0011$ \\
 & Count-matched, overlap & $0.1000 \pm 0.0004$ & $0.1263 \pm 0.0021$ & $0.1180 \pm 0.0013$ & $0.0175 \pm 0.0007$ & $0.1217 \pm 0.0048$ & $0.0194 \pm 0.0007$ \\
\cmidrule(lr){2-8}
 & Overlap, no interaction & $0.0988 \pm 0.0006$ & $0.1081 \pm 0.0013$ & $0.1689 \pm 0.0084$ & $0.0126 \pm 0.0010$ & $0.2035 \pm 0.0128$ & $0.0164 \pm 0.0026$ \\
 & Overlap, with interaction & $0.1016 \pm 0.0004$ & $0.1160 \pm 0.0001$ & $0.1791 \pm 0.0092$ & $0.0246 \pm 0.0013$ & $0.2208 \pm 0.0189$ & $0.0339 \pm 0.0039$ \\
 & Overlap, RMS-matched & $0.0995 \pm 0.0006$ & $0.1104 \pm 0.0024$ & $0.1746 \pm 0.0090$ & $0.0129 \pm 0.0011$ & $0.2105 \pm 0.0130$ & $0.0167 \pm 0.0034$ \\
\bottomrule
\end{tabular}
\end{adjustbox}
\end{table}
\FloatBarrier

Under state-dependent injection (Table~\ref{tab:state-dependent-recovery}), Loki reduces future-effect MAE relative to the zero-response reference by 70\% on Ride-Hailing and 79\% on M5, compared with 93\% and 94\% under stable responses (\textcolor{black}{\hypersetup{linkcolor=black}Tables~\ref{tab:recovery-reference} and~\ref{tab:controlled-event-recovery}}). Thus, Loki still recovers much of the event effect, but recovery becomes less accurate when response intensity depends on a state unavailable to the ERF. We deliberately exclude recent series levels from the ERF inputs to limit compensation for backbone errors in routine dynamics; this also limits its capacity to represent state-dependent responses.
\paragraph{Overlapping events and interactions.}
We first test overlap without an added interaction by keeping event count, type, duration, dose, and response amplitude fixed and shifting event times. For Ride-Hailing, the separated reference is the stable multiplicative setting in Table~\ref{tab:controlled-event-recovery}. For M5, paired schedules place the same two events apart or together within each 28-step block. In the count-matched comparisons in Table~\ref{tab:overlap-interaction-recovery}, future-effect MAE rises from 0.0228 to 0.0261 on Ride-Hailing and from 0.0177 to 0.0194 on M5. Recovery error increases only slightly when overlapping responses follow the multiplicative composition.

With identical event schedules and individual log responses $S_1^*$ and $S_2^*$, the no-interaction series is $Y^{\mathrm{ref}}\exp(S_1^*+S_2^*)$. The interaction condition adds
\begin{equation}
I^*=0.5\min\bigl(|S_1^*|,|S_2^*|\bigr),
\qquad
Y=Y^{\mathrm{ref}}\exp(S_1^*+S_2^*+I^*).
\label{eq:app-interaction-injection}
\end{equation}
This additional term is nonzero only where both responses are nonzero.

To control perturbation strength, we scale the no-interaction log response to match the interaction condition's training-set RMS of standardized target increments, keeping the factor fixed for validation and testing.

Table~\ref{tab:overlap-interaction-recovery} shows that adding the explicit interaction increases future-effect MAE on both datasets, even after matching training perturbation magnitude.

\endgroup
\FloatBarrier

\clearpage
\section{Comparison with Related Methods}
\label{app:method-comparison}

Table~\ref{tab:method-comparison} compares the 19 evaluated models, plus PHR and CycleNet for mechanism comparison only. General-purpose baselines are grouped by input availability; methods with explicit decomposition or correction are listed separately. Parameter sharing refers to sharing between historical decomposition and forecast reconstruction.

\begin{table}[!htb]
\centering
\caption{Input handling, decomposition, and learning mechanisms of the compared methods.}
\label{tab:method-comparison}
\footnotesize
\setlength{\tabcolsep}{3pt}
\renewcommand{\arraystretch}{1.0}
\begin{adjustbox}{max totalheight=\dimexpr\textheight-70pt\relax,max width=\linewidth}
\begin{tabular}{@{}>{\raggedright\arraybackslash}p{\dimexpr.23\textwidth-6pt\relax}>{\raggedright\arraybackslash}p{\dimexpr.22\textwidth-6pt\relax}>{\raggedright\arraybackslash}p{\dimexpr.20\textwidth-6pt\relax}>{\raggedright\arraybackslash}p{\dimexpr.16\textwidth-6pt\relax}>{\raggedright\arraybackslash}p{\dimexpr.19\textwidth-6pt\relax}@{}}
\toprule
Method & Historical decomposition & Forecast reconstruction & Parameter sharing & Parameter learning \\
\midrule
Pure endogenous (DLinear, PatchTST, SparseTSF, TimeMixer, iTransformer) & No explicit event decomposition; raw history as input & Direct forecast & Not applicable & Joint training \\
\addlinespace[1pt]
Historical exogenous (TimeXer, CrossLinear, XLinear) & No explicit event decomposition; events as historical covariates & Direct forecast & Not applicable & Joint training \\
\addlinespace[1pt]
Future exogenous (TiDE, DAG, TFT) & No explicit event decomposition; events as historical and future covariates & Direct forecast conditioned on future covariates & Not applicable & Joint training \\
\addlinespace[1pt]
PHR \citep{wolters2021promotion} & Replace event-week sales with nearby weighted means & Seasonal forecast $\rightarrow$ multiplicative demand model & Not shared & Staged estimation \\
\addlinespace[1pt]
SPADE \citep{wolff2024spade} & Replace event peaks by forward filling & Regular forecast + peak forecast & Not shared & Fixed filling rule; jointly trained prediction network \\
\addlinespace[1pt]
CEDAR \citep{meng2026cedar} & Not decomposed & Backbone forecast + event residual & Not shared & Two-stage training; backbone frozen in stage two \\
\addlinespace[1pt]
AutoARIMA-X (dynamic regression) \citep{hyndman2008autoarima} & Subtract exogenous regression terms, including events; model the remaining series with ARIMA & Exogenous regression forecast + ARIMA forecast & Exogenous regression coefficients & Joint estimation per series \\
\addlinespace[1pt]
SCAN*PRO-style log-linear regression \citep{wittink1988scanpro} & Fit seasonal and exogenous terms, including events, jointly in log space & Seasonal level $\times$ exogenous multipliers & Regression coefficients & Ridge regression per series \\
\addlinespace[1pt]
Prophet \citep{taylor2018prophet} & Fit trend, seasonal, holiday and regressor components & Sum of component forecasts & Component parameters & Joint estimation per series \\
\addlinespace[1pt]
NeuralProphet \citep{triebe2021neuralprophet} & Subtract trend, seasonal, event and regressor components & Autoregressive-network forecast + future components & Component parameters & Joint training of components and autoregressive network \\
\addlinespace[1pt]
NBEATSx \citep{olivares2023nbeatsx} & Subtract backcast components block by block; covariates enter through an exogenous basis & Sum of trend, seasonal and exogenous block forecasts & Block network; separate backcast and forecast coefficients & Joint training of backcast and forecast branches \\
\addlinespace[1pt]
CycleNet \citep{lin2024cyclenet} & Subtract the periodic template & Remaining-series forecast + periodic template & Periodic template & Joint training of template and backbone \\
\addlinespace[1pt]
\textbf{Loki} & \textbf{Divide out the event response; forecast the adjusted series} & \textbf{Adjusted-series forecast $\times$ event response} & \textbf{Event response function (ERF)} & \textbf{Joint training of ERF and backbone} \\
\addlinespace[1pt]
\bottomrule
\end{tabular}
\end{adjustbox}

\smallskip
\raggedright
The NeuralProphet row describes the \href{https://github.com/ourownstory/neural_prophet/blob/0.8.0/neuralprophet/time_net.py\#L577-L638}{official implementation}.
\end{table}
\FloatBarrier

%% file: table/benchmark.tex
\caption{Statistics of the five benchmark datasets. Observations denote entity--time rows.}
\label{tab:dataset-overview}
\centering
\begingroup
\footnotesize
\begin{tabular}{lccccc}
\toprule
Dataset & Ride-hailing & TaxiBJ & M5 & Rossmann & Seoul Bike \\
\midrule
Resolution & 30 min & 30 min & Daily & Daily & Hourly \\
\# Observations & 14,025,600 & 21,872,640 & 137,829 & 1,017,209 & 8,760 \\
\# Entities & 200 & 1,024 & 70 & 1,115 & 1 \\
\# Targets & 9 & 2 & 1 & 1 & 1 \\
\bottomrule
\end{tabular}
\endgroup

%% file: table/appendix_rmse.tex
\begin{table}[!htb]
\caption{Standardized RMSE across five datasets. Results are reported as mean $\pm$ sample standard deviation across seeds. \textcolor{black}{AutoARIMA-X, Log-linear, and Prophet are deterministic fits and are reported without a seed standard deviation.} The best results are highlighted in \textbf{bold} and the second best are \underline{underlined}.}
\label{tab:appendix-rmse}
\centering
\scriptsize
\setlength{\tabcolsep}{2pt}
\begin{adjustbox}{max width=\textwidth}
\begin{tabular}{l|ccccc}
\toprule
Model & \textcolor{black}{Ride-Hailing} & TaxiBJ & M5 & Rossmann & Seoul Bike \\
\midrule
DLinear & $0.4766 \pm 0.0007$ & $0.2372 \pm 0.0009$ & $0.1757 \pm 0.0051$ & $0.5198 \pm 0.0007$ & $0.9116 \pm 0.0167$ \\
PatchTST & $0.4660 \pm 0.0004$ & $0.2642 \pm 0.0065$ & $0.1725 \pm 0.0036$ & $0.4548 \pm 0.0020$ & $0.8846 \pm 0.0087$ \\
SparseTSF & $0.4759 \pm 0.0004$ & $0.2538 \pm 0.0077$ & $0.1725 \pm 0.0034$ & $0.4939 \pm 0.0008$ & $0.8085 \pm 0.0202$ \\
TimeMixer & $0.4703 \pm 0.0015$ & $0.2602 \pm 0.0055$ & $0.1742 \pm 0.0089$ & $0.4493 \pm 0.0067$ & $0.8172 \pm 0.0091$ \\
iTransformer & $0.4698 \pm 0.0013$ & $0.2585 \pm 0.0090$ & $0.1770 \pm 0.0075$ & $0.4503 \pm 0.0014$ & $0.8631 \pm 0.0077$ \\
TimeXer & $0.4725 \pm 0.0007$ & $0.2473 \pm 0.0106$ & $0.1788 \pm 0.0112$ & $0.4492 \pm 0.0062$ & $0.8282 \pm 0.0073$ \\
CrossLinear & $0.4714 \pm 0.0016$ & $0.2488 \pm 0.0046$ & \textcolor{black}{$0.1711 \pm 0.0138$} & $0.4513 \pm 0.0012$ & $0.8487 \pm 0.0125$ \\
XLinear & $0.4683 \pm 0.0009$ & $0.2688 \pm 0.0090$ & $0.1741 \pm 0.0065$ & $0.4442 \pm 0.0062$ & $0.8401 \pm 0.0020$ \\
TiDE & \underline{$0.4481 \pm 0.0023$} & $0.2443 \pm 0.0062$ & $0.1758 \pm 0.0069$ & $0.2449 \pm 0.0037$ & $0.8669 \pm 0.1137$ \\
DAG & $0.4689 \pm 0.0009$ & \textcolor{black}{$0.2508 \pm 0.0055$} & \textcolor{black}{$0.1779 \pm 0.0068$} & \textcolor{black}{\underline{$0.2347 \pm 0.0038$}} & \textcolor{black}{$0.7611 \pm 0.0377$} \\
TFT & $0.4596 \pm 0.0056$ & \textcolor{black}{$0.2317 \pm 0.0110$} & $0.1825 \pm 0.0128$ & $0.2586 \pm 0.0079$ & $\boldsymbol{0.4548 \pm 0.0315}$ \\
\textcolor{black}{AutoARIMA-X} & \textcolor{black}{$5.4221$} & \textcolor{black}{$0.3253$} & \textcolor{black}{$0.2102$} & \textcolor{black}{$0.2784$} & \textcolor{black}{$0.7710$} \\
\textcolor{black}{Log-linear (SCAN*PRO)} & \textcolor{black}{$0.4964$} & \textcolor{black}{\underline{$0.2244$}} & \textcolor{black}{$0.2697$} & \textcolor{black}{$0.2831$} & \textcolor{black}{$0.5348$} \\
\textcolor{black}{Prophet} & \textcolor{black}{$0.7792$} & \textcolor{black}{$0.3587$} & \textcolor{black}{$0.1825$} & \textcolor{black}{$0.2839$} & \textcolor{black}{$1.0445$} \\
\textcolor{black}{NeuralProphet} & \textcolor{black}{$0.4887 \pm 0.0041$} & \textcolor{black}{$0.2558 \pm 0.0153$} & \textcolor{black}{\underline{$0.1693 \pm 0.0053$}} & \textcolor{black}{$0.3065 \pm 0.0150$} & \textcolor{black}{$1.7447 \pm 0.4329$} \\
\textcolor{black}{NBEATSx} & \textcolor{black}{$0.4760 \pm 0.0071$} & \textcolor{black}{$0.2485 \pm 0.0083$} & \textcolor{black}{$0.1746 \pm 0.0133$} & \textcolor{black}{$0.2663 \pm 0.0049$} & \textcolor{black}{$0.7711 \pm 0.1295$} \\
\textcolor{black}{SPADE} & \textcolor{black}{$0.4791 \pm 0.0050$} & \textcolor{black}{$0.2504 \pm 0.0050$} & \textcolor{black}{$0.1961 \pm 0.0193$} & \textcolor{black}{$0.2736 \pm 0.0072$} & \textcolor{black}{$0.6491 \pm 0.0527$} \\
\textcolor{black}{CEDAR} & \textcolor{black}{$0.5437 \pm 0.0470$} & \textcolor{black}{$0.2521 \pm 0.0093$} & \textcolor{black}{$0.1883 \pm 0.0071$} & \textcolor{black}{$0.2504 \pm 0.0132$} & \textcolor{black}{$0.5291 \pm 0.1366$} \\
\midrule
Loki & \textcolor{black}{$\boldsymbol{0.4448 \pm 0.0006}$} & \textcolor{black}{$\boldsymbol{0.2179 \pm 0.0027}$} & \textcolor{black}{$\boldsymbol{0.1582 \pm 0.0021}$} & \textcolor{black}{$\boldsymbol{0.2177 \pm 0.0046}$} & \textcolor{black}{\underline{$0.4969 \pm 0.0232$}} \\
\bottomrule
\end{tabular}
\end{adjustbox}
\end{table}

%% file: table/appendix_wmape.tex
\begin{table}[!htb]
\caption{Original-scale WMAPE across five datasets. Results are reported as mean $\pm$ sample standard deviation across seeds. \textcolor{black}{AutoARIMA-X, Log-linear, and Prophet are deterministic fits and are reported without a seed standard deviation.} The best results are highlighted in \textbf{bold} and the second best are \underline{underlined}.}
\label{tab:appendix-wmape}
\centering
\scriptsize
\setlength{\tabcolsep}{2pt}
\begin{adjustbox}{max width=\textwidth}
\begin{tabular}{l|ccccc}
\toprule
Model & \textcolor{black}{Ride-Hailing} & TaxiBJ & M5 & Rossmann & Seoul Bike \\
\midrule
DLinear & $0.1258 \pm 0.0010$ & $0.1796 \pm 0.0007$ & $0.1027 \pm 0.0026$ & $0.2145 \pm 0.0003$ & $0.5647 \pm 0.0142$ \\
PatchTST & $0.1083 \pm 0.0003$ & $0.1880 \pm 0.0033$ & $0.0982 \pm 0.0013$ & $0.1588 \pm 0.0023$ & $0.4488 \pm 0.0041$ \\
SparseTSF & $0.1216 \pm 0.0002$ & $0.1928 \pm 0.0094$ & $0.0986 \pm 0.0021$ & $0.1762 \pm 0.0006$ & $0.4718 \pm 0.0185$ \\
TimeMixer & $0.1110 \pm 0.0002$ & $0.1862 \pm 0.0031$ & $0.0985 \pm 0.0040$ & $0.1594 \pm 0.0041$ & $0.4239 \pm 0.0054$ \\
iTransformer & $0.1120 \pm 0.0003$ & $0.1861 \pm 0.0053$ & $0.0996 \pm 0.0031$ & $0.1604 \pm 0.0007$ & $0.4659 \pm 0.0066$ \\
TimeXer & $0.1143 \pm 0.0004$ & $0.1881 \pm 0.0070$ & $0.1026 \pm 0.0055$ & $0.1618 \pm 0.0018$ & $0.4371 \pm 0.0101$ \\
CrossLinear & $0.1106 \pm 0.0003$ & $0.1813 \pm 0.0026$ & $0.0983 \pm 0.0074$ & $0.1632 \pm 0.0016$ & $0.4242 \pm 0.0058$ \\
XLinear & $0.1111 \pm 0.0003$ & $0.1975 \pm 0.0081$ & \underline{$0.0979 \pm 0.0035$} & $0.1611 \pm 0.0033$ & $0.4282 \pm 0.0020$ \\
TiDE & \underline{$0.1081 \pm 0.0013$} & $0.1894 \pm 0.0081$ & $0.1038 \pm 0.0044$ & $0.1022 \pm 0.0027$ & $0.5106 \pm 0.0581$ \\
DAG & $0.1107 \pm 0.0008$ & \textcolor{black}{$0.1841 \pm 0.0033$} & \textcolor{black}{$0.1018 \pm 0.0034$} & \textcolor{black}{$0.0994 \pm 0.0013$} & \textcolor{black}{$0.3987 \pm 0.0187$} \\
TFT & $0.1117 \pm 0.0012$ & \underline{$0.1691 \pm 0.0065$} & $0.1025 \pm 0.0059$ & $0.1018 \pm 0.0041$ & \underline{$0.2780 \pm 0.0160$} \\
\textcolor{black}{AutoARIMA-X} & \textcolor{black}{$0.3018$} & \textcolor{black}{$0.2440$} & \textcolor{black}{$0.1133$} & \textcolor{black}{$0.1233$} & \textcolor{black}{$0.4908$} \\
\textcolor{black}{Log-linear (SCAN*PRO)} & \textcolor{black}{$0.1374$} & \textcolor{black}{$0.1724$} & \textcolor{black}{$0.1393$} & \textcolor{black}{$0.1138$} & \textcolor{black}{$0.3011$} \\
\textcolor{black}{Prophet} & \textcolor{black}{$0.1769$} & \textcolor{black}{$0.2897$} & \textcolor{black}{$0.1041$} & \textcolor{black}{$0.1229$} & \textcolor{black}{$0.6498$} \\
\textcolor{black}{NeuralProphet} & \textcolor{black}{$0.1259 \pm 0.0031$} & \textcolor{black}{$0.1879 \pm 0.0078$} & \textcolor{black}{$0.0992 \pm 0.0030$} & \textcolor{black}{$0.1308 \pm 0.0101$} & \textcolor{black}{$1.0392 \pm 0.2218$} \\
\textcolor{black}{NBEATSx} & \textcolor{black}{$0.1155 \pm 0.0003$} & \textcolor{black}{$0.1821 \pm 0.0060$} & \textcolor{black}{$0.1018 \pm 0.0067$} & \textcolor{black}{$0.1131 \pm 0.0025$} & \textcolor{black}{$0.4334 \pm 0.0431$} \\
\textcolor{black}{SPADE} & \textcolor{black}{$0.1207 \pm 0.0009$} & \textcolor{black}{$0.1869 \pm 0.0038$} & \textcolor{black}{$0.1127 \pm 0.0108$} & \textcolor{black}{$0.1100 \pm 0.0035$} & \textcolor{black}{$0.3777 \pm 0.0267$} \\
\textcolor{black}{CEDAR} & \textcolor{black}{$0.1519 \pm 0.0173$} & \textcolor{black}{$0.1771 \pm 0.0062$} & \textcolor{black}{$0.1050 \pm 0.0030$} & \textcolor{black}{\underline{$0.0989 \pm 0.0048$}} & \textcolor{black}{$0.3005 \pm 0.0862$} \\
\midrule
Loki & \textcolor{black}{$\boldsymbol{0.1045 \pm 0.0002}$} & \textcolor{black}{$\boldsymbol{0.1602 \pm 0.0023}$} & \textcolor{black}{$\boldsymbol{0.0921 \pm 0.0010}$} & \textcolor{black}{$\boldsymbol{0.0873 \pm 0.0018}$} & \textcolor{black}{$\boldsymbol{0.2720 \pm 0.0175}$} \\
\bottomrule
\end{tabular}
\end{adjustbox}
\end{table}

%% file: table/appendix_event_regime.tex
\begin{table}[!htbp]
\caption{Indicator-defined event-regime results on Ride-hailing. Standardized MAE values are reported as mean $\pm$ sample standard deviation across \textcolor{black}{five seeds (2027--2031)}. Lower MAE is better.}
\label{tab:appendix-event-regime}
\centering
\begingroup
\footnotesize
\setlength{\tabcolsep}{1.5pt}
\renewcommand{\arraystretch}{1.15}
\begin{tabular*}{\textwidth}{@{\extracolsep{\fill}}lrcccc@{}}
\toprule
Regime & \# Samples & Backbone & History-only & Future-only & \textbf{Full Loki} \\
\midrule
Q$\rightarrow$Q & 4,492 & \textcolor{black}{$0.14887 \pm 0.00016$} & \textcolor{black}{$0.15358 \pm 0.01121$} & \textcolor{black}{$0.14676 \pm 0.00024$} & \textcolor{black}{$\boldsymbol{0.14626 \pm 0.00027}$} \\
E$\rightarrow$Q & 15,793 & \textcolor{black}{$0.14803 \pm 0.00035$} & \textcolor{black}{$0.15313 \pm 0.01332$} & \textcolor{black}{$0.14570 \pm 0.00017$} & \textcolor{black}{$\boldsymbol{0.14451 \pm 0.00031}$} \\
Q$\rightarrow$E & 1,269 & \textcolor{black}{$0.15521 \pm 0.00029$} & \textcolor{black}{$0.15960 \pm 0.01081$} & \textcolor{black}{$\boldsymbol{0.14944 \pm 0.00054}$} & \textcolor{black}{$0.14995 \pm 0.00063$} \\
E$\rightarrow$E & 14,046 & \textcolor{black}{$0.18666 \pm 0.00027$} & \textcolor{black}{$0.19006 \pm 0.01513$} & \textcolor{black}{$0.17563 \pm 0.00030$} & \textcolor{black}{$\boldsymbol{0.17500 \pm 0.00036}$} \\
\bottomrule
\end{tabular*}
\endgroup
\end{table}

%% file: table/event.tex
\begin{table}[htbp]
\centering
\scriptsize
\color{black}
\setlength{\tabcolsep}{3pt}
\caption{\textcolor{black}{Recovery and prediction under multiplicative, additive, and null injections. All entries are standardized MAE, reported as means $\pm$ sample standard deviations across generated configurations after averaging training seeds within each configuration. Bold means indicate the lowest MAE within each dataset and injection condition with multiple models. ``Support'' denotes the artificial response support used for recovery evaluation.}}
\label{tab:controlled-event-recovery}
\begin{adjustbox}{max width=\textwidth}
\begin{tabular}{lllcccc}
\toprule
Dataset & Injection & Model & Prediction MAE & Prediction MAE & Historical MAE & Future effect \\
 & & & (overall) & (support) & after adjustment & MAE \\
\midrule
\multirow{7}{*}{Ride-Hailing} & \multirow{4}{*}{Multiplicative} & Full Loki & $\mathbf{0.1616}\pm0.0003$ & $\mathbf{0.1694}\pm0.0015$ & $\mathbf{0.0207}\pm0.0004$ & $\mathbf{0.0228}\pm0.0007$ \\
 &  & Independent ERFs & $0.1621\pm0.0004$ & $0.1700\pm0.0016$ & $0.0294\pm0.0004$ & $0.0245\pm0.0006$ \\
 &  & Half-independent & $0.1624\pm0.0004$ & $0.1705\pm0.0017$ & $0.0299\pm0.0001$ & $0.0252\pm0.0008$ \\
 &  & Additive Loki & $0.1769\pm0.0003$ & $0.2119\pm0.0013$ & $0.1149\pm0.0019$ & $0.1135\pm0.0015$ \\
\cmidrule(lr){2-7}
 & \multirow{2}{*}{Additive} & Full Loki & $0.1688\pm0.0002$ & $0.1869\pm0.0011$ & $0.0782\pm0.0013$ & $0.1004\pm0.0026$ \\
 &  & Additive Loki & $\mathbf{0.1597}\pm0.0001$ & $\mathbf{0.1606}\pm0.0005$ & $\mathbf{0.0267}\pm0.0005$ & $\mathbf{0.0247}\pm0.0004$ \\
\cmidrule(lr){2-7}
 & Null & Full Loki & $0.1569\pm0.0001$ & $0.1555\pm0.0005$ & $0.0097\pm0.0012$ & $0.0082\pm0.0010$ \\
\midrule
\multirow{7}{*}{M5} & \multirow{4}{*}{Multiplicative} & Full Loki & $\mathbf{0.0996}\pm0.0007$ & $0.1128\pm0.0026$ & $\mathbf{0.0118}\pm0.0004$ & $\mathbf{0.0155}\pm0.0034$ \\
 &  & Independent ERFs & $0.1045\pm0.0008$ & $0.1184\pm0.0032$ & $0.0272\pm0.0031$ & $0.0173\pm0.0035$ \\
 &  & Half-independent & $0.1020\pm0.0004$ & $\mathbf{0.1118}\pm0.0026$ & $0.0277\pm0.0024$ & $0.0174\pm0.0017$ \\
 &  & Additive Loki & $0.1049\pm0.0038$ & $0.1256\pm0.0088$ & $0.0469\pm0.0032$ & $0.0675\pm0.0089$ \\
\cmidrule(lr){2-7}
 & \multirow{2}{*}{Additive} & Full Loki & $0.1000\pm0.0005$ & $0.1191\pm0.0032$ & $0.0319\pm0.0009$ & $0.0552\pm0.0037$ \\
 &  & Additive Loki & $\mathbf{0.0974}\pm0.0012$ & $\mathbf{0.1010}\pm0.0025$ & $\mathbf{0.0134}\pm0.0018$ & $\mathbf{0.0141}\pm0.0027$ \\
\cmidrule(lr){2-7}
 & Null & Full Loki & $0.0964\pm0.0007$ & $0.1021\pm0.0015$ & $0.0052\pm0.0009$ & $0.0062\pm0.0003$ \\
\bottomrule
\end{tabular}
\end{adjustbox}
\end{table}